\documentclass[11pt, a4paper]{article}

\usepackage[utf8]{inputenc}             
\usepackage[T1]{fontenc}              
\usepackage{graphicx}                  
\usepackage{amsmath, amssymb, amsthm}   
\usepackage{hyperref}                  
\usepackage[margin=1in]{geometry}       
\usepackage{enumitem}                 
\usepackage{booktabs}                  
\usepackage[backend=biber, style=numeric-comp, sorting=none]{biblatex}
\usepackage{adjustbox}

\usepackage{amsmath,amssymb,amsfonts}
\usepackage{algorithm}
\usepackage{algpseudocode}
\usepackage{multirow}
\usepackage{graphicx}
\usepackage{textcomp}
\usepackage{color}
\usepackage{url}
\usepackage{booktabs}
\usepackage{multirow}
\usepackage{siunitx}
\usepackage{bm}

\title{A Smartphone-Based Method for Assessing Tomato Nutrient Status through Trichome Density Measurement} 

\author{
SHO UEDA\textsuperscript{1}\thanks{Corresponding author : Sho Ueda} \qquad  XUJUN YE\textsuperscript{2} \\[1em]
\small \textsuperscript{1}United Graduate School of Agricultural Sciences, Iwate University, Morioka,\\ \small Iwate, 020-8550, Japan. \\
\small \textsuperscript{2}Faculty of Agriculture and Life Science, Hirosaki University, Hirosaki, \\ \small Aomori, 036-8560, Japan.
}

\date{\empty} 

\begin{document}

\maketitle

\begin{abstract}
Early detection of fertilizer-induced stress in tomato plants is crucial for optimizing crop yield through timely management interventions.
While conventional optical methods struggle to detect fertilizer stress in young leaves, these leaves contain valuable diagnostic information through their microscopic hair-like structures, particularly trichomes, which existing approaches have overlooked.
This study introduces a smartphone-based noninvasive technique that leverages mobile computing and digital imaging capabilities to quantify trichome density on young leaves with superior detection latency.
Our method uniquely combines augmented reality technology with image processing algorithms to analyze trichomes transferred onto specialized measurement paper.
A robust automated pipeline processes these images through region extraction, perspective transformation, and illumination correction to precisely quantify trichome density.
Validation experiments on hydroponically grown tomatoes under varying fertilizer conditions demonstrated the method's effectiveness. 
Leave-one-out cross-validation revealed strong predictive performance with the area under the precision-recall curve (PR-AUC: 0.82) and area under the receiver operating characteristic curve (ROC-AUC: 0.64), while the predicted and observed trichome densities exhibited high correlation ($r = 0.79$).
This innovative approach transforms smartphones into precise diagnostic tools for plant nutrition assessment, offering a practical, cost-effective solution for precision agriculture.
\end{abstract}

\textbf{Keywords:} Tomato, tomato leaf, fertilizer stress, trichome, smartphone, computer vision, machine learning, precision agriculture, digital agriculture 


\section{Introduction}
\label{sec:introduction}
Tomato (\textit{Solanum lycopersicum}), a vegetable that is widely cultivated worldwide, is influenced by various environmental factors that affect its production and quality.
The nutrient status of plants, which is one of the most crucial factors influencing tomato yield, is an important consideration in tomato production.
Therefore, accurately assessing plant nutrient status is essential for maintaining high tomato yields.

The assessment of plant nutrient status has evolved from invasive to noninvasive approaches.
Initially, techniques were developed to analyze the constituents of fertilizers in soil, media, and nutrient solutions, aiming to estimate the amount of nutrients absorbed by plants \cite{sonneveld1990estimating,chen2015crop}.
Additionally, researchers have developed invasive methods to analyze plant nutrients \cite{hochmuth1994efficiency,andersen1999relationships}.
More recently, noninvasive techniques, including visual inspection and sensor-based monitoring, have become available for assessing plant traits, offering the advantage of continuous monitoring without damaging the plant.

Among these noninvasive techniques, morphological observation has proven particularly valuable for assessing plant responses to environmental stressors, serving as valuable outcome measures of plant stress responses.
Studies have demonstrated strong relationships between morphological changes and various stress conditions, particularly in the context of environmental stressors such as drought and nutrient deficiencies.
Under drought stress, Klepper et al. conducted an observational study and reported that cotton stem diameter, leaf water potential, and relative leaf water content are closely related to the net radiation above the plant canopy \cite{klepper1971stem}.
Wakamori et al. optically analyzed the wilting of tomato leaves and changes in stem diameter in an observational study and demonstrated a correlation between these parameters \cite{wakamori2019optical}.
In terms of fertilizer stress, Padilla et al. provided a comprehensive review of techniques for monitoring the nitrogen status of vegetable crops via optical methods \cite{padilla2018proximal}.
Locascio et al. and Coltman demonstrated a causal relationship between the fertilization rate and tomato yield in interventional studies \cite{locascio1997nitrogen,coltman1988yields}.
Tei et al. reported a causal relationship between the fertilization rate and tomato leaf area in an interventional study \cite{tei2002critical}.
These findings collectively demonstrate how environmental stressors manifest in observable changes in plant morphology.

The ability to detect and examine these modifications allows for a more sophisticated understanding of plant responses to varying environmental conditions.
Building on these morphological indicators, optical measurement techniques have been developed to provide quantitative assessments of plant stress responses.
These methods focus particularly on leaf characteristics such as area and chlorophyll content, which are commonly used as indicators of the adaptive response of plants to fertilizer stress and can be measured through leaf color and spectral analysis.
dela Torre et al. reviewed various optical indices, leaf area, biomass, and other phenotypes for estimating rice yield via remote sensing \cite{delatorre2021remote}.
Cartelat et al. demonstrated that optically assessed contents of leaf polyphenols and chlorophyll in wheat can be used as indicators of nitrogen deficiency \cite{cartelat2005optically}.
Padilla et al. reported that chlorophyll meters and canopy reflectance sensors serve as sensitive indicators of N status across a wide range of vegetable crops \cite{padilla2020monitoring}.
These methods have led to the development of a diagnostic framework based on the relationships among fertilizer application, phenotypic expression, and yield outcomes.
Color and spectral data analysis of plant canopies makes it possible to predict plant health and productivity.

Current optical observation techniques utilize various platforms, from handheld devices to satellites and drones, combined with data collected from in-field sampling stations.
These techniques, which utilize reflected light from the plant body, enable noninvasive and digital monitoring of plant phenotypes \cite{padilla2018proximal}.
The detection of fertilizer stress in tomatoes has been extensively researched via these approaches.
Tables \ref{table:comparison_methods} and \ref{table:measurement_sites} present a comprehensive comparison of various noninvasive methods for the early detection of fertilizer stress in tomato plants.
We focused on studies that employed interventional experiments with varying fertilizer levels and used noninvasive measurement techniques on tomatoes.
Priority was given to methods that utilized digital measurement tools and enabled measurements as close to the growth point as possible for early stress detection.
However, the effectiveness of these methods varies considerably in their practical application.
The selected phenotypic indicator and sensor type influence multiple aspects of measurement.
These aspects include the complexity of data processing for visualization, modeling, and interpretation; the necessary prerequisites during measurement; and practical constraints such as measurement time and operational complexity.

A critical consideration in tomato nutrient assessment is detection latency, the delay between stress onset and detection.
Evaluating phenotypic indicators based on the metabolic mechanisms of tomato plants is crucial, as tomatoes exhibit apical growth, where cell differentiation occurs at the top of the plant and differentiated organs move downward as the plant grows.
Considering this mechanism, measuring phenotypic indicators near the shoot apical meristem could theoretically provide earlier and more accurate detection of fertilizer stress.
The measurement of young leaves should therefore be recognized as a critical nonfunctional requirement for early stress detection.

However, as shown in Table \ref{table:measurement_sites}, existing studies primarily utilize "fully expanded leaves" or even the entire plant as a phenotypic indicator for chlorophyll meters, visible-near infrared spectroscopy, fluorescence spectroscopy, stem diameter measurement, and leaf area measurement.
A "fully expanded leaf" is typically considered the 3rd to 5th compound leaf from the apex, excluding the apical meristem and leaf primordia, and has reached its final size and shape.
Nevertheless, there is no universally accepted definition of a "fully expanded leaf," and its specific characteristics, including the precise distance from the apical meristem, may vary depending on factors such as plant species, cultivar, growth stage, and environmental conditions.
Therefore, it is essential to consider the context of each study when interpreting the term "fully expanded leaf".

This approach may delay stress detection, highlighting a significant gap in early stress detection capabilities.
There is a lack of studies identifying the phenomenon of fertilizer stress, particularly in the assessment of young tomato leaves, with the definition of "young leaves" itself being subject to variation.
Although young leaf measurement is a critical nonfunctional requirement, current methodologies have not adequately addressed this need.
To address these limitations and develop more effective methods for early stress detection, a systematic review of existing measurement approaches and their limitations is necessary, particularly with a focus on the challenges specific to tomato plants.

\begin{table*}[t]
  \footnotesize
  \caption{Comparison of Noninvasive Methods for Detection of Fertilizer Stress in Tomato Plants}
  \label{table:comparison_methods}
  \centering
  \begin{adjustbox}{width=\textwidth}
  \begin{tabular}{lllll}
  \toprule
  \textbf{Method} & \textbf{Phenotypic Indicator} & \textbf{Instrumentation} & \textbf{Conditions to Avoid} & \textbf{Data Complexity} \\
  \midrule
  \textbf{Proposed Method} & Trichome density & Digital camera & Direct sunlight & 2D image array \\
  Leaf Chlorophyll Content \cite{padilla2015threshold} & Chlorophyll a and b content & Chlorophyll meter & Direct sunlight & Scalar value \\
  Visible-Near Infrared Spectroscopy \cite{padilla2015threshold} & Multiple biochemical components & VIS-NIR spectrometer & Direct sunlight & 1D spectral array \\
  Fluorescence Spectroscopy \cite{bahadur2015gas} & Multiple fluorescent compounds & Fluorescence spectrometer & Bright light & 1D spectral array \\
  Leaf Area \cite{xiao2022relationship} & Leaf area & Ruler & Withering & Scalar value \\
  Stem Diameter \cite{olagunju2024effect} & Stem diameter & Caliper & Withering & Scalar value \\
  \bottomrule
  \end{tabular}
  \end{adjustbox}
  \begin{flushleft}
  
  \footnotesize{* This table includes studies that met the following criteria: noninvasive measurement, tested on tomatoes, and interventional experiments with varying fertilizer levels. Priority was given to methods with digital measurements and those enabling measurements near the growth point for early stress detection.}
  
  \end{flushleft}
\end{table*}

\begin{table*}[t]
  \footnotesize
  \caption{Measurement Sites and Sensitivity for Early Detection}
  \label{table:measurement_sites}
  \centering
  \begin{adjustbox}{width=\textwidth}
  \begin{tabular}{lp{8cm}l}
  \toprule
  \textbf{Method} & \textbf{Leaf Age or Plant Part for Early Detection} & \textbf{Sensitivity to Fertilizer Stress} \\
  \midrule
  \textbf{Proposed Method} & \SI{15}{\centi\meter} from shoot meristem (not fully expanded leaf)  & $r = 0.79$, $\text{mPR} = 0.82$, $\text{mROC} = 0.64$ \\
  Leaf Chlorophyll Content \cite{padilla2015threshold} & Fully expanded leaf\textsuperscript{a}  & Correlated\textsuperscript{b} \\
  Visible-Near Infrared Spectroscopy \cite{padilla2015threshold} & Stem area \SI{34}{\centi\meter} below fully expanded leaf\textsuperscript{a}  & Correlated\textsuperscript{b} \\
  Fluorescence Spectroscopy \cite{bahadur2015gas} & Fully expanded leaf\textsuperscript{a}  & $p < 0.05, \,n = \text{not}\,\text{specified}$\textsuperscript{c} \\
  Leaf Area \cite{xiao2022relationship} & All leaves on the plant & Correlated\textsuperscript{b} \\
  Stem Diameter \cite{olagunju2024effect} & The canopy height on the main stem\textsuperscript{a} & $p < 0.05, \,n = 72$\textsuperscript{c} \\
  \bottomrule
  \end{tabular}
  \end{adjustbox}
  \begin{flushleft}
  
  \footnotesize{\textsuperscript{a} Measurement location was not specified in the original study.}\\
  \footnotesize{\textsuperscript{b} Sensitivity to fertilizer stress was not quantitatively assessed in the original study; however, a correlation was visually observed in the presented figures.}\\
  \footnotesize{\textsuperscript{c} Specific sensitivity indicator was not provided in the original study.}\\
  \footnotesize{\textsuperscript{d} mPR and mROC represent the mean area under the precision-recall curve and the mean area under the receiver operating characteristic curve, respectively (see equations \ref{eq:mroc} and \ref{eq:mpr}). These metrics evaluate the performance of the model in determining the need for additional fertilization.}
  
  \end{flushleft}
\end{table*}

\section{Related Work}
To address the challenges identified in early stress detection, particularly in young tomato leaves, this section reviews optical measurements that have emerged as promising approaches for detecting fertilizer stress.
In our literature review, we specifically focused on studies that examined plant responses under varying fertilizer conditions, whether through controlled interventions or through observations across fields with distinct fertilization regimes.
We distinguished these studies from studies conducted under uniform fertilizer conditions that only identified correlations between chemical analyses and plant phenotypes.

Previous studies have explored various optical methods for assessing plant nutrient status.
Of particular interest were studies that combined controlled fertilizer treatments with noninvasive optical assessment methods, as they provide insights into both measurement methodology and its practical limitations.
For example, Gianquinto et al. conducted an intervention study in which a multispectral radiometer positioned directly above the tomato plant canopy was used to investigate the correlations among light reflectance, chlorophyll concentration, and total Kjeldahl nitrogen content.
By employing this noncontact arrangement, measurements were conducted without physically interacting with the leaves.
They discovered a linear relationship between the optical reflectance and chlorophyll concentration; however, no such linear relationship between the optical reflectance and total Kjeldahl nitrogen content was identified \cite{gianquinto2011a}.
Similarly, Padilla et al. directly observed the most recently fully expanded leaves via an optical leaf-clip chlorophyll meter (Konica Minolta, SPAD-502).
A linear relationship was observed between optical observations and the nitrogen nutrition index under conditions where the leaf surface was mechanically flattened by the measurement clips \cite{padilla2015threshold}.
In both investigations, the absence of any barriers along the path of light between the observing device and the plant leaves was assumed, highlighting a gap between the specifications of diagnostic equipment for farmers and the current state of research and development.

\begin{figure}[t!]
  \centering
  \includegraphics[width=0.5\columnwidth, keepaspectratio=true]{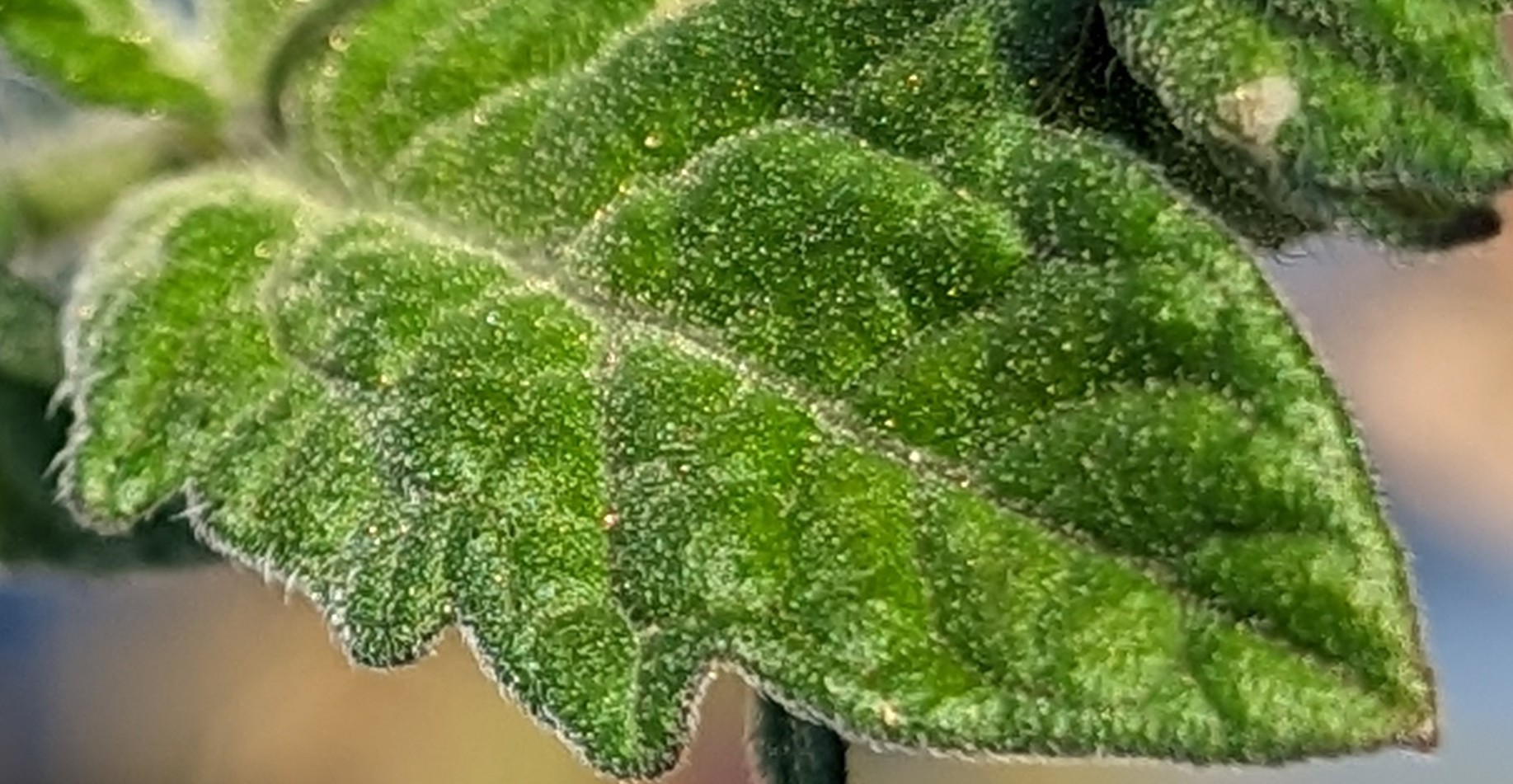}
  \caption{\textbf{Many trichomes-small projections or hair-like structures-are present on the surface of tomato leaves.}}
  \label{fig1}
\end{figure}

While these optical methods have shown promise in various crops, tomato leaves present unique characteristics that create both challenges and opportunities for nutrient status assessment.
Unlike many other crops, tomatoes have numerous trichomes, which are small projections or hair-like structures, on the surface of their leaves (Figure \ref{fig1}).
These trichomes are densely concentrated on young, actively growing leaves and sparsely distributed on older leaves \cite{johnson1975plant}.
The presence of such surface microstructures is known to considerably affect their appearance \cite{kinoshita2005structural,middleton2024selfassembled,thomas2010function}.
Given their high density on young tomato leaves, it is plausible that trichomes significantly influence the effectiveness of optical sensing techniques used to assess plant nutrient status.
Current methodologies do not adequately address this impact when evaluating fertilizer stress, particularly in young leaves.
Therefore, the presence of trichomes should be recognized as an overlooked nonfunctional requirement in measurement device design, and it must be incorporated into the design process.

\begin{figure}[t!]
  \centering
  \includegraphics[width=0.5\columnwidth, keepaspectratio=true]{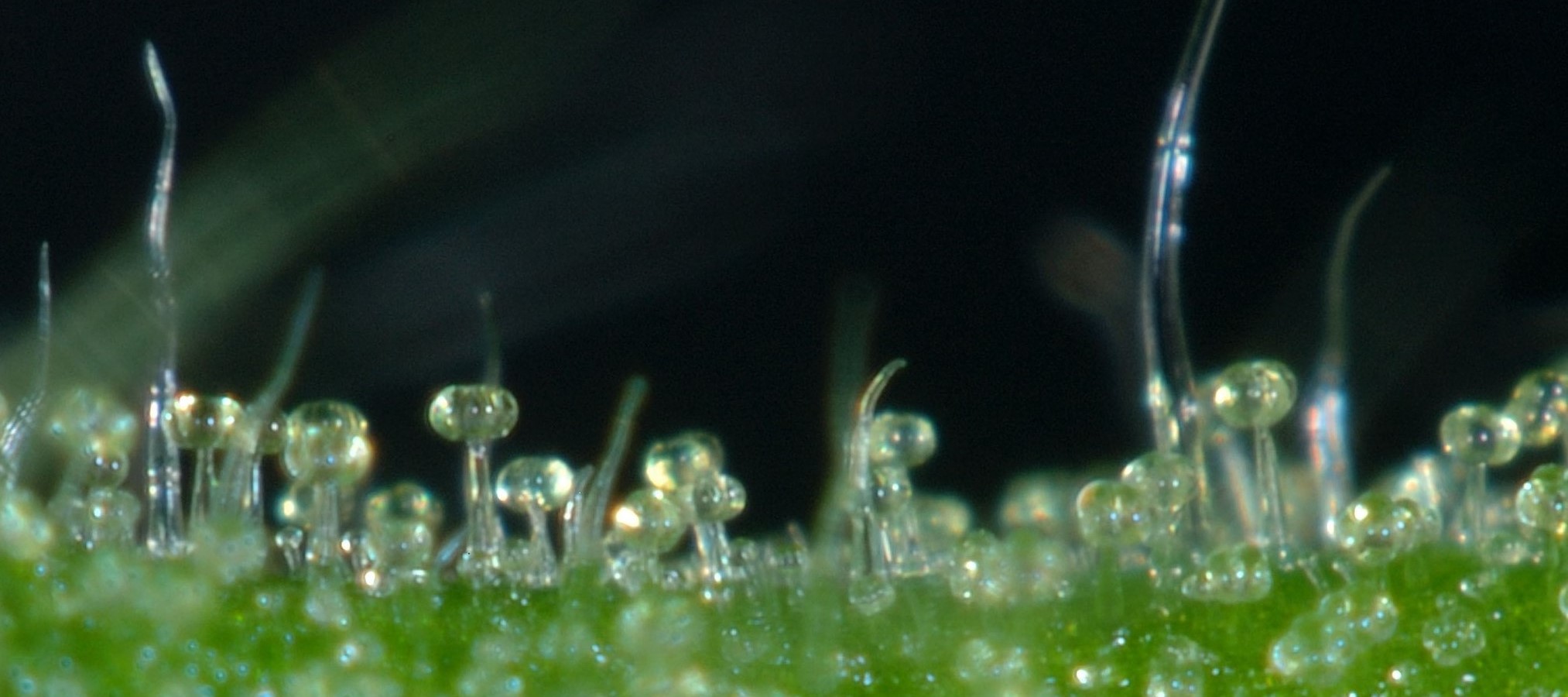}
  \caption{\textbf{Type VI trichomes on the tomato plant surface.}}
  \label{fig2}
\end{figure}

However, trichomes themselves may offer a novel approach for stress detection.
In addition to posing a measurement challenge, trichomes have been identified as potential indicators of fertilizer stress, as they contribute to plant defense against pests by responding to fertilizer stress and altering plant density \cite{barbour1991interaction,hoffland2000nitrogen}.
Various types of trichomes can be observed on the surface of tomato leaves, most of which are short, hair-like trichomes and trichomes with large, spherical cells at their tips (known as type VI trichomes; Figure \ref{fig2}) \cite{wilkens1996resource}.
The glandular head of type VI trichomes is connected to the stalk cell at the base by fragile cell walls \cite{bergau2015the}. This structural arrangement allows the glandular part to easily detach from the leaf surface upon physical stimulation.
The mucilage within the glandular part is known to inhibit pest activity.
This detachment mechanism, along with the mucilage that inhibits pest activity, demonstrates how trichomes actively respond to environmental conditions.
While chlorophyll content, which is correlated with leaf color, remains a widely used indicator of fertilizer stress in plants, the structural and functional characteristics of trichomes suggest their potential as alternative stress indicators.

Multiple studies have demonstrated relationships between trichomes and nutrient status.
Researchers have observed multiple instances where trichomes respond to fertilizer stress \cite{barbour1991interaction,hoffland2000nitrogen,li2014the}.
For example, Barbour et al. revealed an inverse relationship between fertilizer application, leaf nitrogen content, and trichome density \cite{barbour1991interaction}.
Similarly, Hoffland et al. reported a positive correlation between the leaf C:N ratio and trichome density \cite{hoffland2000nitrogen}.
In contrast, Leite et al. reported that trichome density increased over time under reduced fertilizer conditions compared to conventional farming methods \cite{leite}.
However, these studies primarily investigated trichome density within a biological context, with limited exploration of practical measurement techniques for agricultural settings \cite{barbour1991interaction,hoffland2000nitrogen,li2014the}.

\begin{figure}[t!]
  \centering
  \includegraphics[width=0.5\columnwidth, keepaspectratio=true]{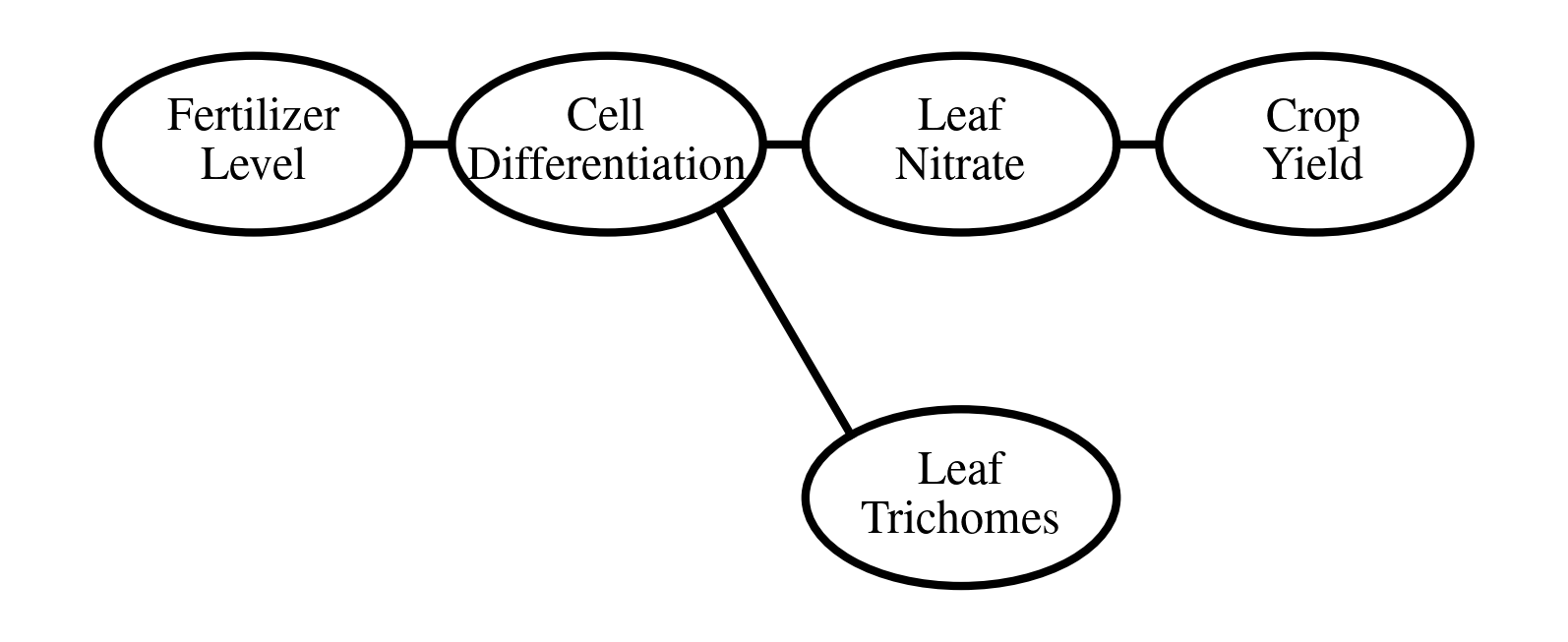}
  \caption{\textbf{Hypothesized relationships between the fertilizer level, cell differentiation, leaf nitrate content, the number of leaf trichomes, and crop yield for tomato plants, serving as the conceptual framework for the study.}}
  \label{fig3}
\end{figure}

Based on these methodological limitations, we identified two potential approaches for overcoming the challenge of early stress detection in young tomato leaves.
These approaches involve modifying existing optical methods through additional data processing and fundamentally redesigning the measurement strategy to utilize trichomes as indicators.
To properly account for trichomes in optical measurements, methods for quantifying their density are essential, regardless of which approach is chosen.
The latter approach is particularly promising because it could satisfy both the prerequisites for optical measurements and the nonfunctional requirement of early stress detection through young leaf measurements with simpler implementation, as it directly utilizes the surface structures that are inherently abundant in young leaves.
As illustrated in Figure \ref{fig3}, our main hypothesis derived from the literature review is that leaf trichome density is influenced by the fertilizer level and could serve as an additional indicator of plant nutrient status, complementing traditional outcome measures such as leaf nitrate content and fruit yield.
To address the gap noted above, the present study focused on tomato leaf trichomes and explored their potential as indicators for assessing the nutrient status of tomato plants in real-world agricultural environments.

To implement this approach, however, practical methods for quantifying trichome density must be developed.
Currently, manual counting via microscopy remains one of the few available methods, and the lack of specialized measurement tools may have hindered progress in this area.
The identification of distinct items within a particular setting containing several objects inside the visual field is commonly considered a computer vision problem \cite{zhao2019object,vincent1991watersheds}.
Given this context, we hypothesized that the use of cameras could transform the challenge of measuring trichome density into an analogy of an object detection problem.

Given these challenges in trichome measurement and the need for practical field-based solutions, we explored the potential of widely available technology.
As highlighted in a comprehensive review by Mendes et al. \cite{mendes2020smartphone}, with their portability, advanced computational capabilities, high-resolution cameras, and internet connectivity, smartphones have enabled various agricultural monitoring applications, including farm management, crop monitoring, and pest and disease diagnosis.
Several successful implementations demonstrate the potential of smartphone-based imaging.
For example, Neumann et al. developed methods for classifying sugar beet leaf diseases \cite{neumann20142014}, Elhassouny et al. obtained similar results with tomato diseases \cite{elhassouny20192019}, and Li et al. demonstrated early detection of tomato late blight via smartphone images of measurement sheets \cite{li2019noninvasive}.
Building on these successes in detecting fine-scale plant features, our study investigated the feasibility of using smartphones as a cost-effective tool for measuring trichome density as an indicator of tomato plant nutrient status.

\section{Materials and Methods}

\subsection{Development of a Smartphone-Based Trichome Density Measurement Method}
\subsubsection{Design of a simple kit for collecting trichomes from tomato leaves}
To establish practical nutrient assessment methods for young tomato leaves, we developed a straightforward measurement kit that addresses the need for young leaf measurements, the limitations of existing optical methods in the presence of trichomes, and the need for field-applicable solutions.
The kit was designed to measure the density of type VI glandular trichomes on tomato leaves as indicators of nutrient status.

\begin{figure}[t!]
  \centering
  \includegraphics[width=0.5\columnwidth, keepaspectratio=true]{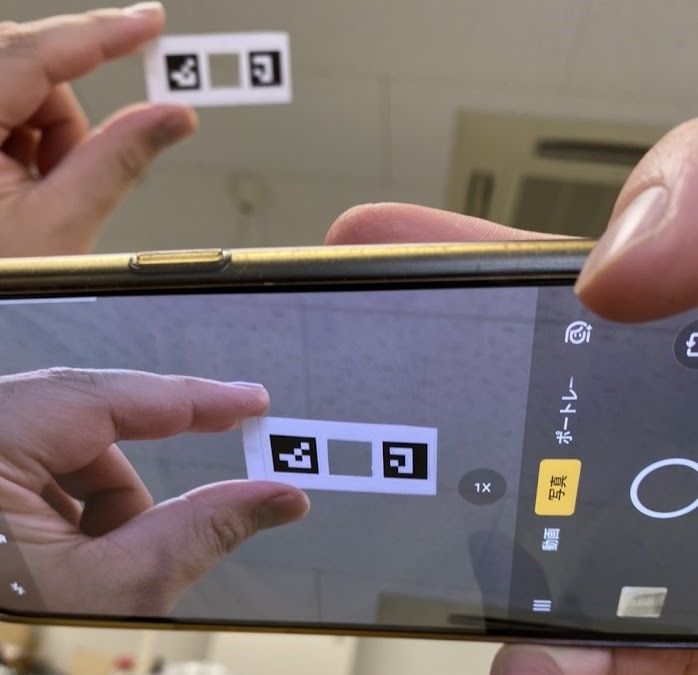}
  \caption{\textbf{A smartphone was used to capture an image of the diagnostic kit.}}
  \label{fig4}
\end{figure}

To ensure accessibility and practical application in agricultural settings, we designed the kit using readily available materials, specifically custom-made measurement paper crafted from inkjet-printable cardstock and cellophane tape (NICHIBAN CT-18).
The selection of cellophane tape was driven by the three-dimensional characteristics of tomato leaves.
The leaf veins and natural leaf curling create macroscopic undulations with longer periods than those caused by trichome surface structures.
Given these topographical features, we needed a measurement method compatible with cameras that have a limited depth of field.
Cellophane tape, which can closely conform to the leaf surface, was selected as both a thin-film substrate and transfer medium, enabling the conversion of three-dimensional trichome distributions into measurable two-dimensional patterns.
This design choice enables cost-effective measurements while maintaining reliability (Figure \ref{fig4}).

To achieve consistent measurements across varying field conditions, the measurement paper incorporates augmented reality (AR) markers, which enable precise distance calibration within captured images, regardless of the camera position or environmental conditions.
Additionally, there is a 12 mm $\times$ 12 mm opening in the center of the paper, which is covered by the cellophane tape for trichome extraction.

The development of this measurement approach was guided by the practical constraints of agricultural applications.
While recent advances in object detection have predominantly relied on complex neural networks requiring high-speed parallel processing, our optimized approach using cellophane tape for trichome extraction circumvents these computational demands.
This technique, combined with simplified image processing algorithms, creates a practical solution that maintains accuracy while eliminating the need for intensive computational resources, which makes the assessment process more economically viable for agricultural applications.

\begin{figure}[t!]
  \centering
  \includegraphics[width=0.5\columnwidth, keepaspectratio=true]{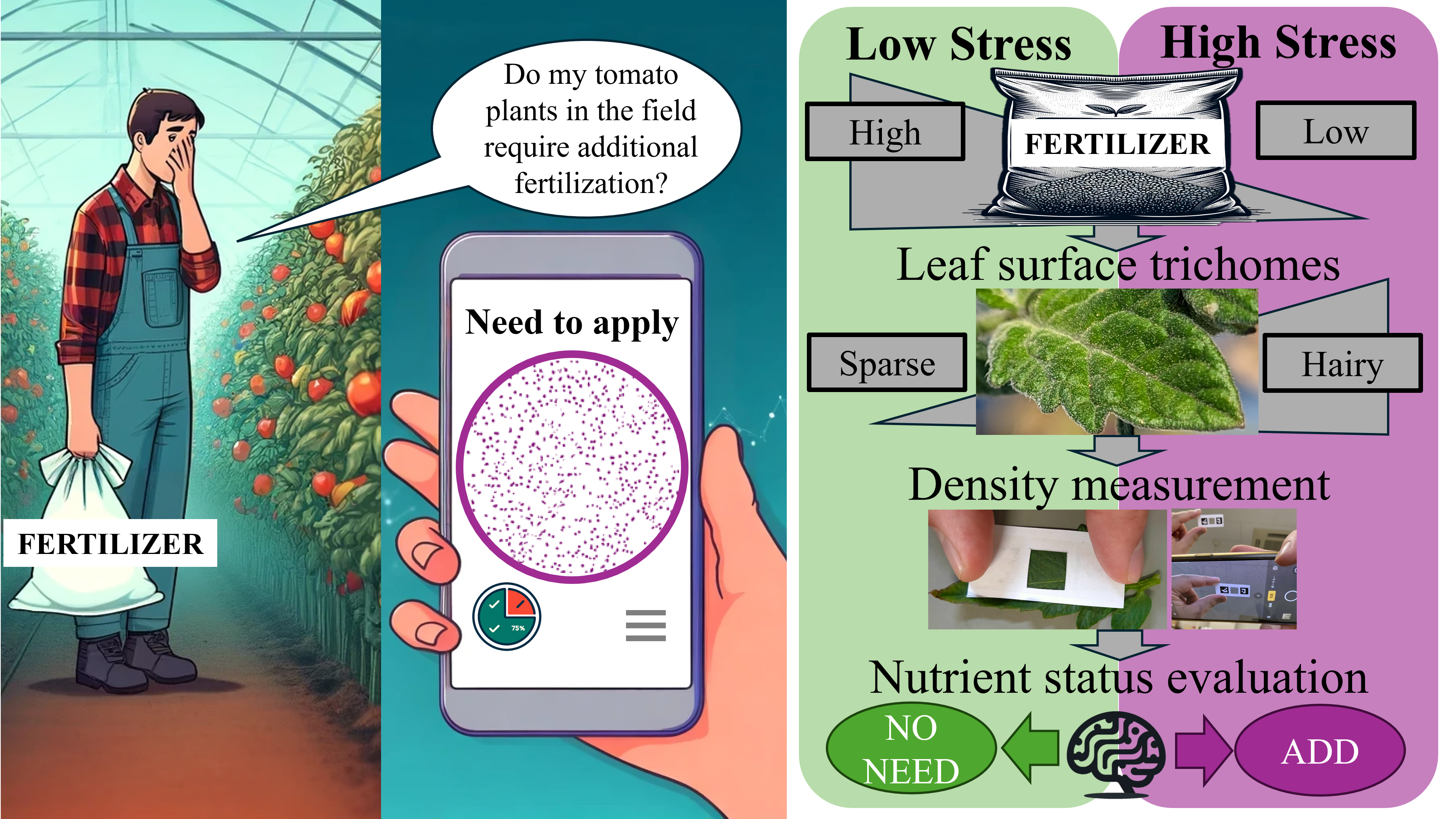}
  \caption{\textbf{Proposed concept of the tomato cultivation support framework.}}
  \label{fig5}
\end{figure}

This diagnostic kit was developed to meet technological needs in agricultural production areas with limited access to laboratory facilities, fulfilling the requirement for reliable nutrient assessment tools in field conditions.
To minimize the number of components required for measurement, the design enables dual manual operation, with the measuring paper in one hand and the smartphone in the other (Figure \ref{fig4}), while maintaining measurement reliability through statistical modeling that accounts for varying optical conditions.
In accordance with lean startup methodology principles \cite{mu2013lean,Ries2011-xn}, the kit was developed as a minimum viable product that prioritizes user-friendly operation without requiring technical expertise, enabling rapid assessment of plant nutrient status in the field (Figure \ref{fig5}).

\subsubsection{Proposed concept of the tomato cultivation support framework}
To translate trichome density measurements into practical agricultural decisions, we developed a systematic support framework that empowers farmers to assess the nutritional needs of their tomato plants.
This framework streamlines the assessment process through four sequential steps.
The process begins with sampling trichomes from tomato leaves via adhesive cellophane tape on measurement paper.
Next, a smartphone is used to capture an image of the transferred trichomes.
The captured image is then analyzed to determine trichome density as an indicator of fertilizer stress.
Finally, specific fertilization recommendations are generated based on the analysis results. This structured approach enables farmers to make informed decisions about whether additional fertilization is necessary, replacing subjective assessments with objective evaluations.

\subsubsection{Overview of the trichome detection image processing pipeline}
The success of this support framework critically depends on accurate and reliable trichome density measurements.
To realize such measurements while maintaining practical utility in agricultural settings, we developed an image processing pipeline that prioritizes computational efficiency and result interpretability.
Rather than employing deep learning models for direct image processing, we chose a hybrid approach that combines traditional computer vision techniques with tabular data analysis.
This architecture leverages the inherent simplicity and reduced dimensionality of tabular data, effectively minimizing computational costs while bolstering quantitative model interpretation.
This approach enables the use of computationally lighter models while improving both the reliability and interpretability of the results for agricultural applications.

Our development process prioritized methodological rigor and practical reliability.
To avoid data leakage, we developed the image processing pipeline using images from preliminary experiments, ensuring that the model development and evaluation processes remained independent and unbiased.

The pipeline consists of three main stages, beginning with preprocessing, where we extract metadata in the exchangeable image file format (EXIF) embedded in the image files.
At this stage, we identify AR markers within the images for initial alignment, apply homography transformations, and adjust for variations in lighting conditions.
Additional preprocessing steps include image binarization for trichome segmentation and noise removal filtering to enhance trichome visibility.

In the feature extraction stage, we focus on identifying trichomes in the images, specifically targeting type VI glandular trichomes owing to their unique morphology.
Using the aligned images and calibrated AR markers from the preprocessing stage, we convert the pixel counts to relative distances, enabling accurate distance estimation.
The final density calculation stage leverages the extracted features to determine trichome density from images captured under various environmental conditions and camera settings, each associated with distinct EXIF data.
The computational process is detailed in Algorithm \ref{alg:trichome_density_analysis}, with the software specifications provided in Table \ref{table:versions}.

\begin{algorithm}
  \caption{Trichome density analysis process}
  \label{alg:trichome_density_analysis}
  \begin{algorithmic}[1]
  \Require $image$: Image file
  \Ensure $density$: Nearest neighbor density
  \Ensure $expTime$: Exposure time from EXIF data
  \Ensure $iso$: ISO speed from EXIF data
  \Ensure $openingPx$: Pixel count of measurement paper opening
  
  \Function{ReadExifAndLoad}{$image$}
      \State Read EXIF data from $image$
      \State $expTime, iso \gets$ Extract from EXIF data
      \State $I(x,y) \gets$ Convert $image$ to the array
      \State \Return $I(x,y)$, $expTime$, $iso$
  \EndFunction
  
  \Function{DetectMarkers}{$I(x,y)$}
      \State Detecting AR markers in $I(x,y)$
      \State $markers \gets$ Extract coordinates from markers
      \State $openingPx \gets$ Calculate the opening pixel count
      \State \Return $openingPx$, $markers$
  \EndFunction
  
  \Function{ApplyHomography}{$I(x,y)$, $markers$}
      \State Apply homography transform to $I(x,y)$
      \State \Return $I_{trans}(x,y)$
  \EndFunction

  \Function{CorrectIllum}{$I(x,y)$}
      \State Apply illumination correction to $I(x,y)$
      \State \Return $I_{corr}(x,y)$
  \EndFunction

  \Function{RemoveNoise}{$I(x,y)$}
      \State Apply noise removal to $I(x,y)$
      \State \Return $I_{clean}(x,y)$
  \EndFunction

  \Function{SegmentTrichomes}{$I(x,y)$}
      \State Apply watershed segmentation to $I(x,y)$
      \State \Return $trichomes$
  \EndFunction
  
  \Function{CalcDensity}{$trichomes$}
      \State $density \gets$ Calculate nearest neighbor density
      \State \Return $density$
  \EndFunction
  \Statex
  \Statex \textbf{Main Execution}
  \State $I(x,y)$, $expTime$, $iso \gets \Call{ReadExifAndLoad}{image}$
  \State $openingPx$, $markers \gets \Call{DetectMarkers}{I(x,y)}$
  \State $I_{trans}(x,y) \gets \Call{ApplyHomography}{I(x,y), markers}$
  \State $I_{corr}(x,y) \gets \Call{CorrectIllum}{I_{trans}(x,y)}$
  \State $I_{clean}(x,y) \gets \Call{RemoveNoise}{I_{corr}(x,y)}$
  \State $trichomes \gets \Call{SegmentTrichomes}{I_{clean}(x,y)}$
  \State $density \gets \Call{CalcDensity}{trichomes}$
  
  \end{algorithmic}
\end{algorithm}

\begin{table}[!t]
\renewcommand{\arraystretch}{1.3}
\caption{Versions of Libraries and Operating System Used}
\label{table:versions}
\centering
\begin{tabular}{ll}
\toprule
\textbf{Operating System/Library} & \textbf{Version} \\
\midrule
Ubuntu & 22.04.2 LTS \\
opencv-contrib-python & 4.6.0.66 \\
imbalanced-learn & 0.12.0 \\
lightgbm & 3.3.5 \\
shap & 0.43.0 \\
ExifRead & 3.0.0 \\
\bottomrule
\end{tabular}
\end{table}

\subsubsection{Image preprocessing: AR marker detection, illumination correction, and noise reduction}
Image processing begins with reading EXIF metadata from image files. This processing provides crucial information about environmental conditions and camera settings during image capture.
Analysis of specific parameters, such as exposure time and ISO speed ratings, enables us to account for variations in imaging conditions that affect trichome density measurements, ensuring consistency across different environments.

Our spatial calibration incorporates AR markers into the image processing pipeline.
These markers serve as reference points for determining the scale and orientation of the captured images, enabling trichome density determination.
For marker detection, we implemented a nonrecursive approach via the ArUco module from the OpenCV library (version 4.6.0.66) \cite{intelcorporationopencvcontribpython}, as the development of AR technology was not the primary focus of this study.

For robust detection under field conditions, we implemented a systematic image preprocessing approach.
The algorithm converts the captured image to grayscale, followed by binarization via Otsu's method to increase marker visibility across lighting conditions.
We selected the "ArUco dictionary" DICT\_4X4\_250 due to its balance between marker uniqueness and detection reliability.
The markers are identified via the OpenCV function cv2.aruco.detectMarkers().
The complete marker detection algorithm is presented in Algorithm \ref{alg:ar_marker_detection}.

\begin{algorithm}
\caption{AR marker detection}
\label{alg:ar_marker_detection}
\begin{algorithmic}[1]
\Require $I$: Input image
\Ensure $C$: Detected marker corners
\Ensure $ids$: Detected marker IDs
\Function{ToGray}{$img$}
\Statex Convert $img$ to grayscale
\Return grayscale image
\EndFunction
\Function{Binarize}{$img$}
\Statex Apply Otsu's binarization
\Return binary image
\EndFunction
\Function{GetDict}{$type$}
\Statex Create ArUco dictionary
\Return dictionary
\EndFunction
\Function{GetParams}{}
\Statex Create detector parameters
\Return parameters
\EndFunction
\Function{DetectM}{$img, dict, params$}
\Statex Detect ArUco markers
\Return corners and IDs
\EndFunction
\Statex
\Statex \textbf{Main Execution}
\State $G \gets \Call{ToGray}{I}$
\State $B \gets \Call{Binarize}{G}$
\State $D \gets \Call{GetDict}{DICT\_4X4\_250}$
\State $P \gets \Call{GetParams}{}$
\State $C, ids \gets \Call{DetectM}{B, D, P}$
\end{algorithmic}
\end{algorithm}

Following AR marker detection, we apply homography transformation to correct perspective distortion in images.
This distortion arises from positional and orientational variations in the measurement paper relative to each camera captured.
The transformation adjusts images as though the camera and measurement paper were parallel, using the coordinates of four corners obtained from AR markers as reference points to estimate a perspective transformation matrix.
This standardization ensures accurate trichome density analysis, regardless of variations in camera positioning during image capture.

Image quality optimization begins with illumination correction, which is an essential step to address nonuniform lighting conditions across the acquired images and to maintain consistent trichome detection despite ambient lighting variations.
For this purpose, we implemented a flat field correction technique \cite{seibert1998medical}.
The process starts by resizing all images to a standardized dimension of 1000 pixels $\times$ 1000 pixels, followed by applying an averaging filter with a kernel size of 50 pixels $\times$ 50 pixels to generate a smoothed version representing the overall illumination pattern.
The correction is mathematically described in Equation \ref{eq:1}.

\begin{equation}
  C = \frac{R \cdot m}{F}
\label{eq:1}
\end{equation}

where $C$ denotes the corrected image, $R$ represents the original (precorrection) image, $m$ is the average luminance of the original image, and $F$ is the smoothed image representing the illumination pattern.
The details of this correction are presented in Algorithm \ref{alg:illumination_correction}.

\begin{algorithm}
\caption{Illumination correction}
\label{alg:illumination_correction}
\begin{algorithmic}[1]
\Require $R$: Precorrection image
\Ensure $C$: Corrected image

\Function{ApplyBlur}{$image, kw, kh$}
    \State $w \gets$ width of $image$
    \State $h \gets$ height of $image$
    \State $output \gets$ new image of size $w \times h$
    \For{$y \gets 0$ to $h - 1$}
        \For{$x \gets 0$ to $w - 1$}
            \State $sum \gets 0$
            \State $count \gets 0$
            \For{$j \gets -kh/2$ to $kh/2$}
                \For{$i \gets -kw/2$ to $kw/2$}
                    \State $nx \gets x + i$
                    \State $ny \gets y + j$
                    \If{$0 \leq nx < w$ and $0 \leq ny < h$}
                        \State $sum \gets sum + image[ny][nx]$
                        \State $count \gets count + 1$
                    \EndIf
                \EndFor
            \EndFor
            \State $output[y][x] \gets sum / count$
        \EndFor
    \EndFor
    \State \Return $output$
\EndFunction

\Function{CalculateAverageLuminance}{$image$}
    \State $sum \gets 0$
    \State $count \gets 0$
    \For{each pixel $p$ in $image$}
        \State $sum \gets sum + p$
        \State $count \gets count + 1$
    \EndFor
    \State \Return $sum / count$
\EndFunction

\Statex
\Statex \textbf{Main Execution}
\State $F \gets \Call{ApplyBlur}{R, 50, 50}$
\State $m \gets \Call{CalculateAverageLuminance}{R}$
\State $C \gets \frac{R \cdot m}{F}$

\end{algorithmic}
\end{algorithm}

The final preprocessing step addresses image noise, which is widely known to reduce detection accuracy in object detection tasks.
We employed morphological opening transformation via a 3 $\times$ 3 square structuring element with all the elements set to 1, applied in a single iteration.
This technique effectively eliminates small-scale noise and artifacts while preserving the overall shape and size of larger objects, and it is especially efficient in reducing background noise.
The structuring element size was chosen to be smaller than the target glandular trichomes, which are approximately \SI{60}{\micro\meter} \cite{bergau2015the} in diameter, ensuring preservation of the features of interest.

\subsubsection{Feature extraction: Trichome identification and density analysis}

For individual trichome identification, we implemented a segmentation procedure based on the watershed algorithm following the noise removal step.
This method detects boundary surfaces that separate contours in the image, offering an efficient approach for rule-based segmentation suited to the distinctive appearance of trichomes.
While deep learning-based segmentation can achieve high accuracy, it typically requires substantial computational resources, including graphics processing units with advanced parallel processing capabilities, to be integrated into the computer system.
In contrast, the watershed algorithm, which constitutes a traditional image-processing technique, operates with lower computational costs on standard computer hardware such as central processing units, making it particularly suitable for achieving our goal of developing accessible and practical tools for agriculture.

The identification process focuses specifically on type VI trichomes present on the adhesive surface of the cellophane tape.
These trichomes exhibit distinctive characteristics: spherical heads \cite{glas2012plant}, a fragile layer facilitating head detachment \cite{bergau2015the}, and a relatively large average diameter of approximately \SI{60}{\micro\meter} \cite{bergau2015the} compared to other trichome types.
Our detection method employs physical and optical filtering on the basis of these specific properties, operating under the assumption of substantial glandular trichome presence on the leaf surface.

In this paper, we utilize the term "trichomes" to describe the contours of detected objects on the basis of their specific properties.
These criteria for identification include the following:

\begin{itemize}
  \item A detachable layer must exist, allowing peeling with cellophane tape;
  \item The object should be larger than a pixel, with a minimum size of approximately \SI{10}{\micro\meter}, and it must be detectable in the image;
  \item The lower limit of the shape factor within the survey area is defined as the 1st quartile minus 1.5 times the interquartile range;
  \item The upper limit of the shape factor within the survey area is defined as the 3rd quartile plus 1.5 times the interquartile range;
  \item An object's shape should more closely resemble a sphere than a prolate ellipsoid, with the circularity exceeding a defined lower limit;
  \item The perimeter of the object falls within a specified range, defined by a lower limit and an upper limit;
  \item The area of the object lies within a specified range, defined by a lower limit and an upper limit.
\end{itemize}

\begin{figure}[t!]
  \centering
  \includegraphics[width=0.5\columnwidth, keepaspectratio=true]{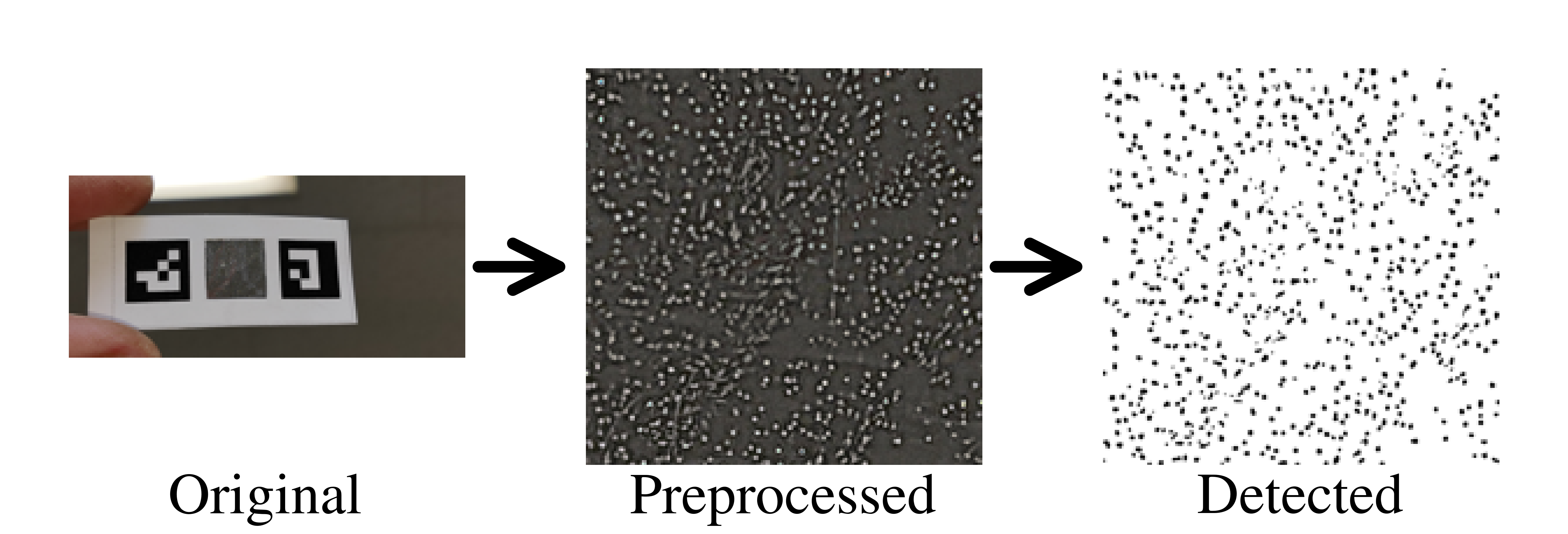}
  \caption{\textbf{Example of image data preprocessing and object detection.}}
  \label{fig6}
\end{figure}

The contours detected through segmentation yield centroid points in two-dimensional space, represented via the pixel coordinates of the image.
Figure \ref{fig6} shows the image processing steps performed thus far, including preprocessing and object detection, leading to these contour detections.
The set of centroids, denoted as $P$, is expressed in Equation \ref{eq:2}.

\begin{equation}
P = \{ \bm{p}_i \in \mathbb{Z}^2 \mid i = 1, 2, \ldots, N \}
\label{eq:2}
\end{equation}

where $\bm{p}_i$ represents the $i$-th centroid point in the two-dimensional integer space $\mathbb{Z}^2$ and $N$ is the total number of detected contours.
Each point $\bm{p}_i$ is represented by its pixel coordinates $(x_i, y_i)$, where $x_i$ and $y_i$ correspond to the horizontal and vertical positions in the image, respectively.

\begin{algorithm}
\caption{Trichome identification}
\label{alg:trichome_identification}
\begin{algorithmic}[1]
\Require $I$: Preprocessed image
\Ensure $P$: Set of trichome centroid points

\Function{IsValidTrichome}{$contour, C_{th}, P_{l}, P_{h}, A_{l}, A_{h}$}
    \State $C \gets \Call{CalculateCircularity}{contour}$
    \State $P \gets \Call{CalculatePerimeter}{contour}$
    \State $A \gets \Call{CalculateArea}{contour}$
    \If{$C < C_{th}$}
        \State \Return false
    \EndIf
    \If{$P < P_{l}$ or $P > P_{h}$}
        \State \Return false
    \EndIf
    \If{$A < A_{l}$ or $A > A_{h}$}
        \State \Return false
    \EndIf
    
    \State \Return true
\EndFunction

\Function{CalculateAllMetrics}{$contours$}
    \State Calculate the circularity, perimeter, and area for all contours
    \State \Return $C_{all}, P_{all}, A_{all}$
\EndFunction

\Function{CalculateCentroidPoint}{$contour$}
    \State Calculate the centroid point $(x, y)$ of the $contour$
    \State \Return $(x, y)$
\EndFunction

\Statex
\Statex \textbf{Main Execution}

\State $contours \gets \Call{WatershedSegmentation}{I}$
\State $P \gets \emptyset$
\State $C_{all}, P_{all}, A_{all} \gets \Call{CalculateAllMetrics}{contours}$
\State $C_{th} \gets 1.5 \times Q1(C_{all})$
\State $P_{l}, P_{h} \gets 1.5 \times Q1(P_{all}), 1.5 \times Q3(P_{all})$
\State $A_{l}, A_{h} \gets 1.5 \times Q1(A_{all}), 1.5 \times Q3(A_{all})$
\For{each $contour$ in $contours$}
    \If{\Call{IsValidTrichome}{$contour, C_{th}, P_{l}, P_{h}, A_{l}, A_{h}$}}
        \State $centroidPoint \gets \Call{CalculateCentroidPoint}{contour}$
        \State $P \gets P \cup \{centroidPoint\}$
    \EndIf
\EndFor

\end{algorithmic}
\end{algorithm}

The detailed process of trichome identification, including the criteria and thresholds used for filtering, is presented in Algorithm \ref{alg:trichome_identification}.
This algorithm encapsulates the key steps and decision-making process for distinguishing type VI glandular trichomes from other objects.

To quantify the spatial distribution of the identified trichomes, we computed the density of the centroid points (denoted as $P$ in Equation \ref{eq:2}) via the nearest neighbor distance (NND) method \cite{clark1954distance}.
This approach measures the distance between each point and its nearest neighbor within the set, as described in Equation \ref{eq:3}.
In this equation, $\mathrm{d}(\bm{p}_i, \bm{q}_i)$ represents the Euclidean distance between point $\bm{p}_i$ and its closest neighboring point $\bm{q}_i$, with the NND calculated by averaging these distances over the total number of contours $N$ present in the image.

\begin{equation}
\mathrm{NND} = \frac{1}{N} \sum_{i=1}^N \mathrm{d}(\bm{p}_i, \bm{q}_i)
\label{eq:3}
\end{equation}

The efficient computation of nearest neighbors becomes crucial when dealing with numerous points.
A brute-force approach, which calculates the distances between all point pairs, has a computational complexity of $O(dn^2)$, where $n$ is the number of data points and $d$ is the dimensionality.
As described by this "big O" notation \cite{skiena1998algorithm}, the execution time of the algorithm increases as the input size ($n, d$) increases, making the brute-force approach computationally expensive as the number of points increases.
To address this computational challenge, we adopted a k-dimensional (k-d) tree algorithm, which organizes points in a space-partitioning data structure for rapid nearest neighbor queries.
This approach effectively reduces the computational complexity to $O(n \cdot d \log n)$ \cite{cunningham2022knearest}, where the execution time increases linearly with dimensionality and increases more moderately with the number of points compared to the brute-force approach.
This logarithmic growth characteristic makes the k-d tree algorithm much more efficient than the brute-force approach, especially for large datasets.
The algorithm is particularly suitable for our two-dimensional image analysis ($d=2$) with numerous trichome centroids.
The detailed implementation of the NND calculation is presented in Algorithm \ref{alg:nnd_analysis}.

\begin{algorithm}
\caption{Density analysis}
\label{alg:nnd_analysis}
\begin{algorithmic}[1]
\Require $P$: Set of trichome centroid points
\Ensure $NND$: nearest neighbor distance
\Function{CalcNND}{$P$}
\State $T \gets \Call{BuildKDTree}{P}$
\State $S \gets 0$
\For{each $p$ in $P$}
\State $q \gets \Call{FindNN}{T, p}$
\State $d \gets \Call{EuclidDist}{p, q}$
\State $S \gets S + d$
\EndFor
\State \Return $S / |P|$
\EndFunction
\Function{BuildKDTree}{$P$}
\State Construct a k-d tree from points $P$
\State \Return tree
\EndFunction
\Function{FindNN}{$T, p$}
\State Find the nearest neighbor of $p$ in tree $T$, excluding $p$
\State \Return nearest neighbor
\EndFunction
\Function{EuclidDist}{$p, q$}
\State $dx \gets p.x - q.x$
\State $dy \gets p.y - q.y$
\State \Return $\sqrt{dx^2 + dy^2}$
\EndFunction
\Statex
\Statex \textbf{Main Execution}
\State $NND \gets \Call{CalcNND}{P}$
\end{algorithmic}
\end{algorithm}

The NND serves as an inverse indicator of spatial density, with higher values denoting sparse arrangements and lower values representing denser groupings.
While conventional density measurements simply count objects per unit area, this measure proves particularly valuable for analyzing type VI glandular trichomes, which are susceptible to detachment from leaf surfaces due to factors such as insect infestations or physical abrasion.
Compared with conventional counting methods, the NND method demonstrates superior robustness to accidental detachment.
The reason for this robustness is that the NND method represents density as a probability distribution, enabling reliable estimates even when localized areas experience trichome detachment, provided that unaffected regions persist.
As a prerequisite for NND calculations, at least two points must be present in the measurement area.
The appendix further explains this concept through a simulation employing point cloud data generated via a Poisson process in two-dimensional space.

\subsection{Evaluation of the New Method for Assessing the Nutrient Status of Tomato Plants}
\subsubsection{Selection of the cultivation method}
A well-controlled cultivation experiment is essential for evaluating our proposed method for assessing tomato plant nutrient status while maintaining testability.
The aim of the experiment was to measure key phenotypes, i.e., trichome density, leaf nitrate ion concentration, and fruit yield, under ideal growth conditions.
We chose hydroponics over soil-based cultivation to avoid the complexities inherent in soil properties, such as variations in composition and structure that can lead to inconsistent water retention, unpredictable fertilizer dynamics, and increased risk of pest and disease outbreaks.
This approach enabled precise control over nutrient availability, and it minimized confounding factors in our evaluation of fertilizer stress while providing an ideal growth environment that approximates the optimal conditions found in commercial cultivation systems.

From the available hydroponic techniques, we selected the deep water culture (DWC) method, where plant roots are fully submerged in nutrient solution within a production tank, which is aerated to sufficiently oxygenate the roots \cite{hoagland1938the,velazquezgonzalez2022a}.
The precise control of nutrient conditions at the individual plant level makes this method particularly suitable for implementation in a randomized block design.
Additionally, its simplicity of setup and minimal equipment requirements facilitate both the relocation and replication of experimental units across locations, enhancing the experiment's falsifiability.

\subsubsection{Tomato growth conditions in hydroponic culture}
We conducted hydroponic cultivation of the tomato cultivar (TAKII SEED, Fruitica) via the DWC method described above in a temperature-controlled greenhouse at Hirosaki University, Japan.
The greenhouse environment was maintained via a heater controller (NEPON, HKC-253) set to \SI{15}{\celsius} and a ventilation controller (Nihon Operator, JRM-102R) set to \SI{25}{\celsius}, with natural light serving as the sole illumination source.
The planting arrangement consisted of fifteen-liter plastic buckets placed in beds measuring 180 cm $\times$ 90 cm, accommodating 8 plants per bed.

The nutrient solution in each bucket was continuously aerated and mixed via a blower system (YASUNAGA, AP-40P), which distributed air through eight segments of silicone tubing and dispersed it into the nutrient solution via air stones for proper oxygenation.
To prevent algae growth in the nutrient solution, the buckets were fitted with lids featuring center holes for tomato stem placement.

\subsubsection{Fertilizer concentration settings}
To investigate the relationship between fertilizer stress and tomato phenotypes, we prepared nutrient solutions using hydroponic fertilizer (OAT Agrio, Fertilizer Series SA) at three concentrations of total dissolved solids (TDS) to induce various levels of fertilizer stress: 1500 ppm (high), 1000 ppm (medium), and 500 ppm (low).
This interventional research design enabled us to examine both the presence of intervention effects on tomato phenotypes and their dose-response relationships.

The selection of these concentrations requires careful consideration because of the lack of standardized protocols for controlling fertilizer levels in DWC tomato cultivation.
While previous studies have reported successful tomato growth in closed hydroponic systems with similar fertilizer compositions at an electrical conductivity (EC) of \SI{1.1}{\deci\siemens\per\metre} \cite{ishihara2007effect}, adopting a single universal value would be inappropriate given the diverse objectives and conditions that individual growers optimize for.
This consideration underscores the importance of our study's focus on detecting fertilizer stress through plant phenotypes.
Using the standard nutrient solution composition provided in the technical documentation of SA fertilizer, we converted this EC value of 1.1 dS$\cdot$m$^{-1}$ to approximately 500 ppm TDS, which became our baseline low-concentration treatment.

We established the medium and high concentrations at two and three times this baseline.
Under our greenhouse conditions, preliminary cultivation experiments confirmed the suitability of this range, showing no visually identifiable growth defects or plant death at 500 ppm and no apparent growth inhibition due to increased osmotic pressure even at 1500 ppm.
These results validated our concentration range selection while supporting our primary goal of detecting fertilizer stress through plant phenotypes to provide actionable fertilization recommendations.

\subsubsection{Experimental design and management}

\begin{figure}[t!]
  \centering
  \includegraphics[width=0.5\columnwidth, keepaspectratio=true]{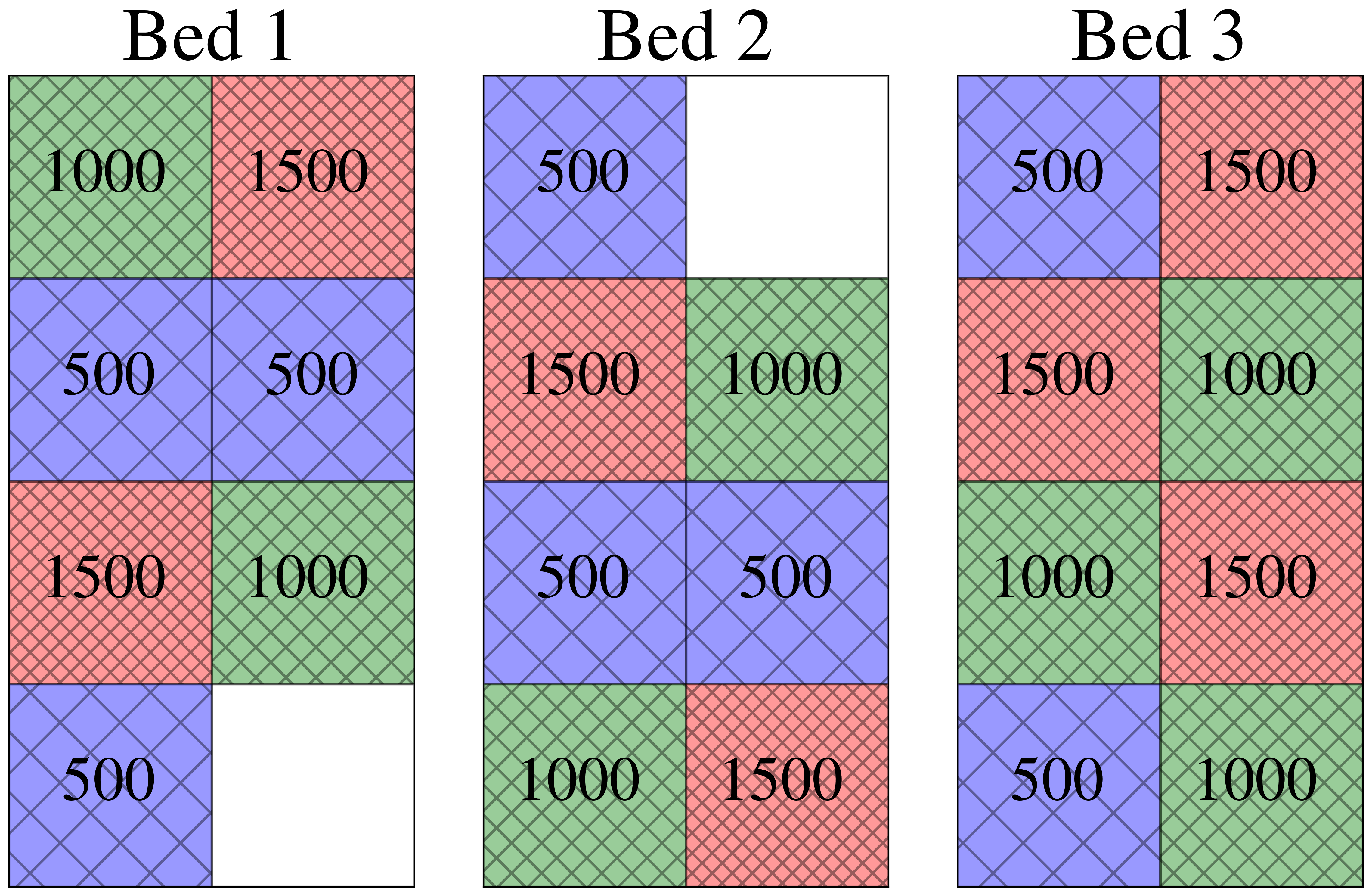}
  \caption{\textbf{Experimental layout with three different fertilizer levels (ppm) in a randomized block design.}}
  \label{fig7}
\end{figure}

To evaluate the effects of fertilizer levels on key outcome variables, i.e., nitrate ion concentration in tomato leaves, trichome density, and yield, we implemented a randomized block design that minimized the potential impact of spatial variations in environmental factors \cite{fisher1966design}.
The fertilizer treatment plots were arranged as shown in Figure \ref{fig7}, with the preliminary experiment utilizing the vacant areas depicted in the figure.

Nutrient solution management followed a systematic protocol throughout the cultivation period.
Water was added to the buckets when nutrient solution levels decreased, while fertilizer was applied biweekly, with concentrations adjusted to the desired TDS levels via a TDS meter (EUTECH, PCSTestr35).
Although the fertilizer composition closely matched the nutrient uptake ratios of tomatoes, enabling operation as a closed hydroponic system without frequent solution replacement \cite{ishihara2007effect}, practical considerations necessitated occasional solution replacement when the solution became cloudy due to algae or root debris.

Plant survival was monitored throughout the experiment, with some attrition occurring at different stages.
Two plants were lost before the compound leaf harvesting stage, reducing the study population to 20 plants.
By the end of the fruit harvesting period, the loss of two additional plants further reduced the population to 18 plants for yield assessment.

\subsubsection{Compound leaf sampling and trichome transfer}
Our experimental protocol included systematic collection of compound leaves from tomato plants, with a total of 40 samples collected across two harvesting sessions in March and April 2023.
To ensure consistency and minimize potential biases, we implemented a structured sampling process where samples were processed sequentially according to ascending plant identification numbers.
These IDs had been randomly assigned to different fertilizer treatment levels during the randomized block design setup, preventing any treatment-related bias in the sampling order.

The sampling location was standardized at 15 cm below the apical meristem of each plant.
When a suitable compound leaf was unavailable at this position, we temporarily skipped the plant to allow for additional growth, continuing the sampling process iteratively until all the plants were sampled.
For each compound leaf, we selected all leaflets larger than the measurement paper opening for trichome density measurements.
The number of measurement papers matched the number of selected leaflets per compound leaf, allowing for accurate assessment of trichome density distribution across each leaflet.

\begin{figure}[t!]
  \centering
  \includegraphics[width=0.5\columnwidth, keepaspectratio=true]{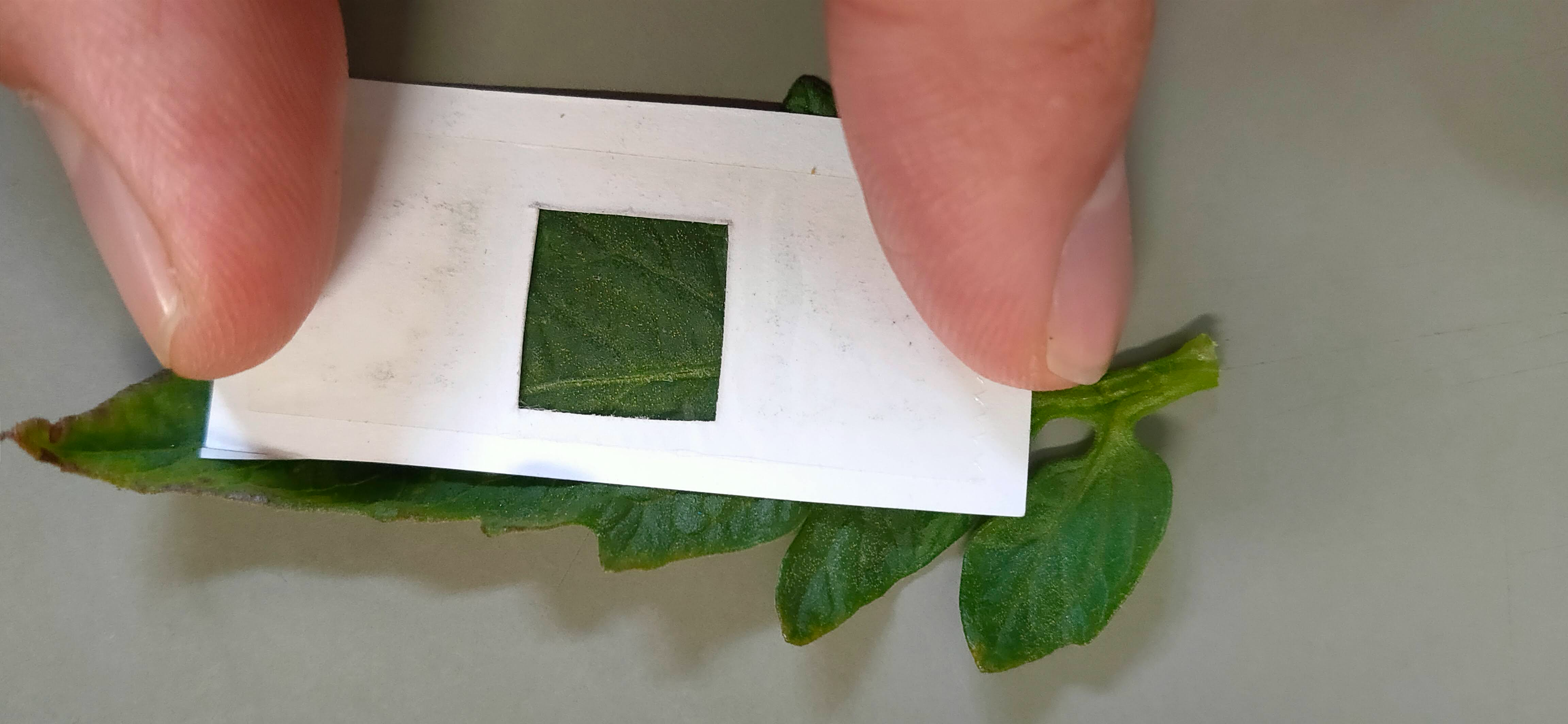}
  \caption{\textbf{Trichome transfer process from the leaflet surface to the measurement paper using cellophane tape.}}
  \label{fig8}
\end{figure}

The glandular trichome transfer process followed a simple yet precise protocol that balanced measurement reliability with practical field applicability.
Each leaflet was carefully positioned on the adhesive side of the cellophane tape (Figure \ref{fig8}) and gently pressed with a finger to achieve complete contact between the leaf surface and the tape.
Following leaflet removal, we photographed the cellophane tape containing the transferred trichomes alongside the measurement paper via a smartphone camera.

\subsubsection{Image acquisition and dataset preparation}
Image acquisition was performed via a smartphone (OPPO, Reno A) equipped with a 16-megapixel camera featuring an F1.7 aperture and a CMOS sensor (Sony, IMX398).
We utilized the device's native camera application with its autofocus capability and automatic adjustment of exposure, ISO sensitivity, and focus parameters, leveraging built-in optimization algorithms to ensure clear images across varying conditions.

To ensure a comprehensive dataset that accounts for various usage scenarios, we manually varied the image capture parameters to validate system robustness.
The distance between the camera sensor and measurement paper ranged from 79 to 218 mm, with horizontal offset angles ranging from -12.52 to 10.62 degrees and vertical offset angles ranging from -9.04 to 8.13 degrees, simulating natural variations in user handling.
The measurement paper was positioned with sufficient background separation to ensure that the background was not clearly identifiable in the captured images.
Photography was conducted in a room with a north-facing window, which eliminated direct sunlight and provided diffuse natural light, supplemented by the smartphone's built-in white LED and oblique ceiling fluorescent lighting (Figure \ref{fig4}).
Photographic sessions were scheduled across different times of day, with solar elevation angles ranging from -7.95 to 33.42 degrees.

Quality control measures were implemented throughout the image acquisition process.
Each image was composed with the measurement paper centered in the frame, and preliminary analysis was conducted immediately after capture to verify pipeline compatibility.
All the measurement papers were carefully checked for dataset inclusion, with immediate rephotography performed when the analysis program encountered exceptions, mostly due to failures in two-dimensional code detection.

The detection accuracy of these codes primarily depends on the code type and detection libraries used.
As two-dimensional code technology has already achieved practical reliability levels, users in agricultural field conditions can simply retake photographs when the measurement software indicates detection exceptions.
Given this independence between code detection and our primary objectives of trichome density measurement and fertilization recommendations, we excluded code detection accuracy from our verification targets.

After the data collection and preprocessing steps, information from all the measurement papers was compiled into the final dataset, which comprised 6,855 images that successfully passed through the image processing pipeline.
This comprehensive dataset encompasses a wide range of distances, angles, and lighting conditions, ensuring that the developed model is trained on and evaluated based on diverse scenarios that closely mimic real-world usage of the kit.

\subsubsection{Nitrate concentration measurement in tomato leaves}
Following trichome density measurements, we developed a systematic process for extracting and analyzing leaf nitrate concentrations.
The tomato leaves were first frozen via a refrigerant (HFO-1234ze) to disrupt cell walls through freezing, facilitating subsequent juice extraction.
The leaves were placed in plastic bags with their zippers partially closed, leaving a small opening for refrigerant introduction.
During the freezing process, the evaporating refrigerant generated dry gas that displaced the moist air from the bags, thus preventing frost formation.

The sample preparation process proceeded through several stages.
The frozen leaves were thawed and mechanically pressed with pliers to extract juice, which was then stored at \SI{-20}{\celsius}.
After storage, the leaf juice was thawed and centrifuged to separate the solid and liquid components.
The clear supernatant was then carefully collected, ensuring that no sediment particles were transferred.
This extract was diluted 20-40 times with electrophoresis buffer (Otsuka Electronics, $\alpha$-AFQ133) and filtered through a \SI{0.45}{\micro\meter} pore size syringe filter.

\begin{figure}[t!]
  \centering
  \includegraphics[width=0.5\columnwidth, keepaspectratio=true]{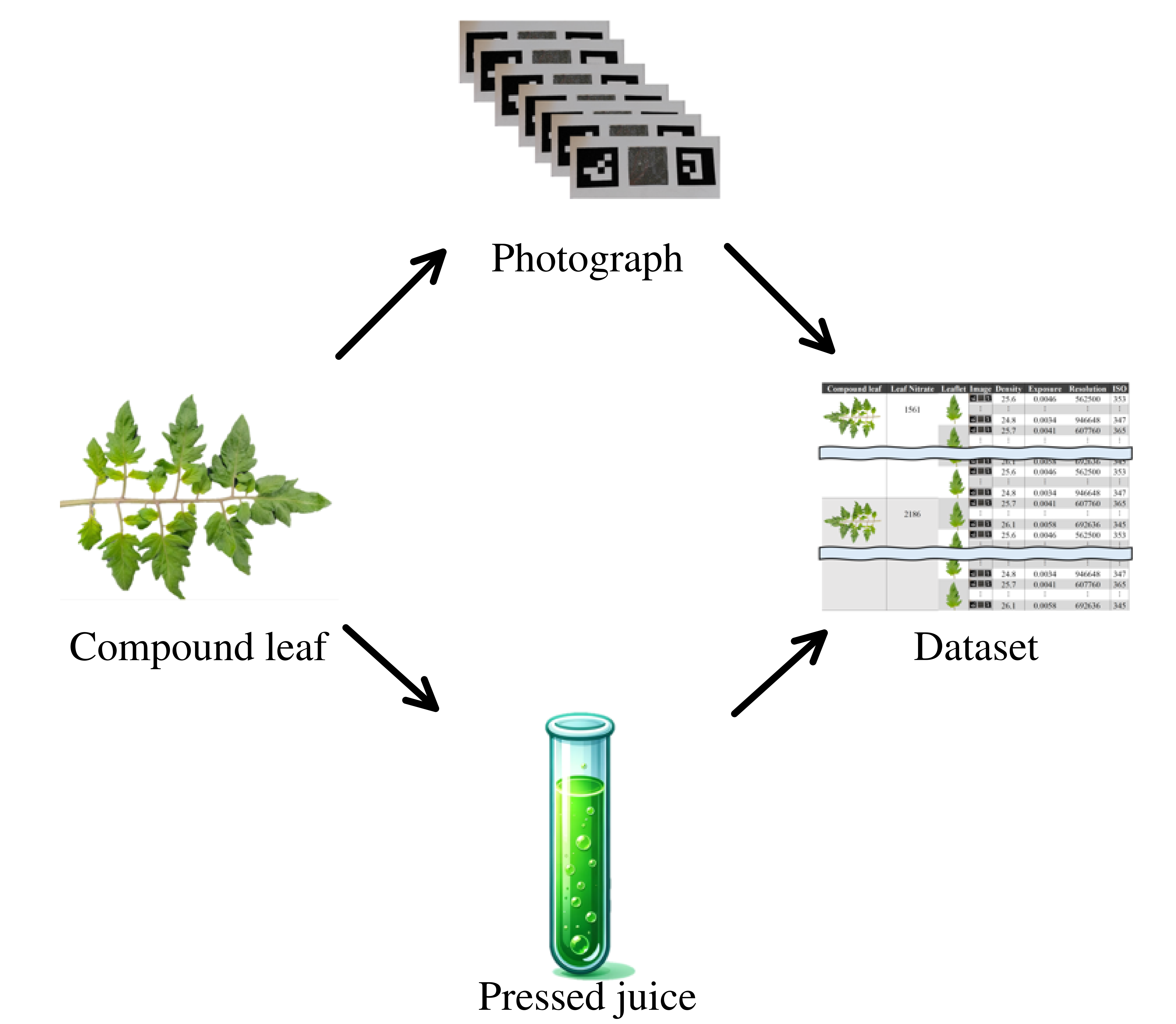}
  \caption{\textbf{Workflow from data acquisition to analysis.}}
  \label{fig9}
\end{figure}

Nitrate ion concentration measurements were performed via a capillary electrophoresis system (Agilent Technologies, 7100) with a capillary tube of \SI{75}{\micro\meter} internal diameter and \SI{72}{\centi\meter} length.
The analysis was conducted under controlled conditions: buffer temperature at \SI{25}{\celsius}, voltage at \SI{20}{\kilo\volt}, injection time of 4 s, injection pressure of \SI{50}{\milli\bar}, and detection wavelength of \SI{350}{\nano\meter}.
The concentrations were determined via absolute calibration against standard samples of known concentration, with one measurement per compound leaf yielding 40 scalar values.
These measurements were integrated with the trichome density data from 6,855 images to create the complete dataset (Figure \ref{fig9}).

\subsubsection{Assessment of tomato fruit harvest under experimental conditions}
The fruit yield of the tomato plants was evaluated under experimental conditions by harvesting all fully ripened red fruits from each plant.
The harvested fruits were stored at \SI{5}{\celsius} and subsequently weighed.

\subsection{Preliminary Data Analysis and Model Preparation}
\subsubsection{Configuration of the data analysis environment}
The analytic environment was built via container virtualization technology to ensure the consistency and reproducibility of the analytical results across different computer systems, regardless of the specific hardware specifications.
The analysis was conducted via Python \cite{pythonsoftwarefoundationpython} version 3.10; Table \ref{table:versions} provides a comprehensive list of the Python library versions and operating system versions utilized in this study.
We used opencv-contrib-python for image processing, imbalanced-learn \cite{scikitlearncontribimbalancedlearn} for dataset preprocessing, lightgbm \cite{kguolinandmicrosoftresearchlightgbm} for machine learning model training, shap \cite{slundbergandsleeshap} for model interpretation, and ExifRead \cite{sianareexifread} for metadata extraction from image data.

\subsubsection{Statistical analysis of nitrate ion concentrations}
To statistically validate the observations of nitrate ion concentrations across different fertilizer levels, we employed a series of nonparametric tests.
First, the Kruskal-Wallis test was used to examine the overall distribution differences among the three fertilizer levels.
Next, pairwise Mann-Whitney U tests were conducted to assess the differences between each pair of fertilizer levels, with Bonferroni correction applied to the resulting p values to account for multiple comparisons.

\subsubsection{Multicollinearity and variable interaction assessment}
Before model construction, we conducted a comprehensive analysis of the relationships among the predictor variables, specifically examining multicollinearity and interaction effects.
Multicollinearity, which occurs when predictor variables are strongly intercorrelated, can lead to unstable and unreliable regression coefficient estimates.
We assessed multicollinearity via the variance inflation factor (VIF), which measures the extent to which the variance of a regression coefficient estimate increases owing to the correlation among predictor variables.
Following common practice, variables with VIF values exceeding 10 were considered to indicate problematic multicollinearity \cite{zuur2010a}, and they were removed from subsequent analyses to enhance model interpretability.

We also examined the interaction effects among the predictor variables, which arise when the level of a predictor variable influences the impact of another predictor variable on the response variable.
Our investigation focused particularly on how image resolution might modulate the relationship between trichome density and leaf nitrate concentration.
This analysis was conducted through stratification, dividing the data into subgroups on the basis of levels of suspected interacting variables and examining the predictor-response relationships within each subgroup.

\subsection{Analysis of Tomato Fertilization Requirements based on Leaf Trichome Density}

\subsubsection{Problem formulation and objectives}
The primary objective of this study was to develop a kit suitable for practical use in the field.
By leveraging existing knowledge about the correlation between trichome density and leaf nitrogen-related indicators, we aimed to address a crucial question that farmers face, which is affected by the fertilizer stress level and trichome density: Do my tomato plants in the field require additional fertilization?

\subsubsection{Model design and variable selection}
We formulated a binary classification problem to determine the need for fertilization on the basis of leaf nitrate ion concentrations.
Given that there is no universally ideal level of fertilizer stress for tomatoes, as optimal levels vary with environmental conditions and cultivator goals, we designed our model with a parameterized nitrate ion concentration threshold.
Using this threshold, our model makes binary decisions.
Concentrations above the threshold indicate that no additional fertilization is needed, whereas concentrations below the threshold suggest that fertilization is needed.

The model used three explanatory variables: NND, image resolution, and exposure time.
These variables underwent standard scaling to ensure comparable scales and to improve model performance.
This scaling technique transformed each variable to have a mean of zero and a standard deviation of one.
For model evaluation, we employed a cross-validation process that divided the data into training and test sets.

The primary focus of this research was to develop a kit suitable for practical field use, emphasizing improvements at the architecture level, that is, the system's fundamental structure and design.
We aimed to develop objective methods for tomato characterization rather than optimizing model performance.
Therefore, we used the model's default hyperparameter values instead of implementing extensive tuning with a validation dataset.
While this decision might lead to underfitting or overfitting due to insufficient control over model complexity, potentially reducing generalization performance and limiting the opportunities for theoretical performance analysis \cite{mohri2018foundations}, we considered these trade-offs to be acceptable, as our priority was creating an accessible and practical tool rather than maximizing theoretical model performance.

\subsubsection{Cross-validation strategy}
Our model validation approach was determined by the hierarchical structure of our dataset.
We analyzed 40 compound leaves harvested from 20 tomato plants across three intervention groups, which yielded 300 leaflets and, ultimately, 6,855 analyzed images.
Given this relatively small-scale but diverse dataset, we employed leave-one-out cross-validation (LOOCV) at the compound leaf level to evaluate the model's generalization performance.
This approach was selected to minimize bias from specific samples while maximizing the use of our limited data points.

LOOCV implementation was supported by our feature extraction process, which enables learning with tabular data while maintaining low computational costs despite the iterative nature of LOOCV.
In each iteration, we selected one compound leaf as the test set and used the data from all remaining compound leaves for training.

\subsubsection{Evaluation methodology}
Our evaluation strategy was designed to reflect the practical usage of our kit, where users can easily perform multiple measurements cost-effectively.
We developed a method to assess model performance by averaging the predictions from multiple images, simulating realistic usage scenarios where farmers might take between 1 and 25 images for a single assessment.
This approach was implemented by randomly extracting 1 to 25 rows of data from the test set and inputting them into the model for prediction.
By examining the model's performance across different numbers of measurements, we could evaluate how prediction accuracy improves with increased measurement intensity.

For a comprehensive performance assessment, we aggregated predictions and actual observations across all cross-validation iterations to calculate various performance metrics.
This aggregated evaluation approach provides insights into the model's reliability under different measurement scenarios while maintaining practical relevance to field applications.

\subsubsection{Parameter range selection}
Our predictive model fundamentally depends on two key parameters that govern its operation.
These parameters are the nitrate ion concentration threshold and the number of images used for evaluation.
To effectively visualize and analyze the results, we carefully defined specific ranges for these parameters, considering both practical utility and model reliability.
This selection was particularly crucial, as it directly impacts our model's ability to provide actionable recommendations under the inherent variability of agricultural conditions.

The determination of suitable nitrate ion concentration thresholds was influenced by two critical factors from our domain expertise.
First, nonlinear models such as light gradient boosting machine (LightGBM) typically produce unstable predictions near or outside their training data boundaries.
Second, agricultural workers and related actors sustain their livelihoods through crop production revenues, where excessive fertilizer stress directly leads to decreased crop productivity and, consequently, reduced profitability.
Based on these considerations, we excluded the low nitrate concentration region from the threshold settings, as such conditions would be impractical and economically unsustainable in real-world agricultural applications.

For the number of evaluation images, we balanced measurement comprehensiveness with practical field constraints.
While our kit enables multiple measurements to be performed easily and cost-effectively, we established an upper limit of 25 images per evaluation.
This limit provides sufficient data for reliable analysis while remaining feasible within the time and resource efficiency requirements.
We defined our primary range of interest for nitrate ion concentrations from the second to third quartiles of our experimental distribution.
For specific visualizations, we adopted a "representative parameter setting" using the third quartile value as the nitrate ion concentration threshold and 25 images for evaluation, providing a practical and statistically robust foundation that reflects real-world agricultural conditions.
Comprehensive performance evaluations across all parameter ranges, including those outside our defined range of interest, are presented in the appendix, along with the detailed data analysis workflow (Algorithm \ref{alg:fertilization_prediction}).

\subsubsection{Feature preprocessing and label setting}
Our modeling objective was to determine the need for additional fertilization by comparing the nitrate ion concentration, estimated from trichome density measurements in images, with a predefined threshold.
We assigned binary labels to leaves based on their nitrate ion concentration: 0 for concentrations below the threshold, indicating no need for additional fertilization, and 1 for concentrations above the threshold, indicating that fertilization was needed.

\subsubsection{Model training}
Prior to modeling, we partitioned our dataset into training and testing sets based on compound leaves for binary classification.
To address the imbalances in both image numbers across compound leaves and leaflets and in class labels, we applied the synthetic minority oversampling technique (SMOTE) \cite{haixiang2017learning,chawla2002smote} to oversample the minority class.

For classification, we employed LightGBM, a decision tree-based approach that combines multiple trees through gradient boosting techniques \cite{ke2017lightgbm}.
In this approach, each decision tree iteratively improves the model's predictive accuracy by learning from the residuals of previous trees.
The regression predictions $z$ were calculated via Equation \ref{eq:4}, where $\mathrm{f}_k(\bm{x})$ represents the output of the $k$-th decision tree, with $K$ denoting the total number of trees and $\bm{x}$ representing the input feature vector.

\begin{equation}
z = \sum_{k=1}^{K} \mathrm{f}_k(\bm{x})
\label{eq:4}
\end{equation}

These predictions were transformed into probabilities through the sigmoid function shown in Equation \ref{eq:5}, which maps values to probabilities between 0 and 1, representing the likelihood of the input belonging to class 1.
This transformation provides farmers with interpretable probabilities of nutrient deficiency, facilitating informed decision making about additional fertilization.
The final classification ($\hat{Y}$) was determined from the computed probability ($\hat{y}$) following the decision rule in Equation \ref{eq:6}.
The model was trained via LightGBM's default parameters, with a learning rate of 0.1 and 100 boosting rounds.

\begin{equation}
\hat{y} = \frac{1}{1 + \mathrm{e}^{-z}}
\label{eq:5}
\end{equation}

\begin{equation}
\hat{Y} = \begin{cases}
1 & \text{if } \hat{y} \geq 0.5 \\
0 & \text{if } \hat{y} < 0.5
\end{cases}
\label{eq:6}
\end{equation}

\subsubsection{Model evaluation and feature contribution}

We evaluated our model by comparing predicted labels $\hat{Y}$ with the true labels via multiple performance metrics.
The area under the precision-recall curve (AUC-PR) served as the primary evaluation metric during training, as defined by Equation \ref{eq:11}.
For comprehensive model assessment based on test data, we employed precision, recall, the $\mathrm{F}_{1} \text{ Score}$, the area under the receiver operating characteristic curve (AUC-ROC), the area under the precision recall curve (AUC-PR), mean AUC-ROC (mROC), and mean AUC-PR (mPR), as described in Equations  \ref{eq:7}-\ref{eq:mpr}.

The foundational metrics include precision (Equation \ref{eq:7}), which measures the ratio of correctly predicted positive observations to total predicted positives \cite{hossin2015a,jeni2013facing}, and recall (Equation \ref{eq:8}), which reflects the model's ability to identify all relevant positive instances.
The $\mathrm{F}_{1} \text{ Score}$ (Equation \ref{eq:9}) provides a balanced measure by calculating the harmonic mean of precision and recall.
For threshold-independent evaluation, we utilized the AUC-ROC (Equation \ref{eq:10}), which summarizes the model's discrimination ability through the relationship between sensitivity (true positive rate, $\mathrm{t}(f)$) and specificity ($1 - \text{false positive rate}$), and the AUC-PR (Equation \ref{eq:11}), which is particularly suitable for imbalanced datasets \cite{jeni2013facing}.

\begin{equation}
\text{Precision} = \frac{\text{True Positive}}{\text{True Positive} + \text{False Positive}}
\label{eq:7}
\end{equation}

\begin{equation}
\text{Recall} = \frac{\text{True Positive}}{\text{True Positive} + \text{False Negative}}
\label{eq:8}
\end{equation}

\begin{equation}
\mathrm{F}_{1} \text{ Score} = 2 \cdot \frac{\text{Precision} \cdot \text{Recall}}{\text{Precision} + \text{Recall}}
\label{eq:9}
\end{equation}

\begin{equation}
\text{AUC-ROC} = \int_0^1 \mathrm{t}(f) \, \mathrm{d}f
\label{eq:10}
\end{equation}

\begin{equation}
\text{AUC-PR} = \int_0^1 \mathrm{p}(r) \, \mathrm{d}r
\label{eq:11}
\end{equation}

To evaluate model performance across different nitrate ion concentration thresholds in operational scenarios, we used two averaged metrics called mROC (Equation \ref{eq:mroc}) and mPR (Equation \ref{eq:mpr}).
For each threshold $c$ within the range of 1600-1900 ppm, we calculate $\text{AUC-ROC}_c$ and $\text{AUC-PR}_c$.
This threshold range was chosen based on observed nitrate ion concentrations in high-yielding tomato leaves, as it would be impractical in agricultural applications to use thresholds associated with decreased crop productivity.
The final metrics are computed by dividing the cumulative values by $M$, where $M$ is the number of unique models generated for each threshold $c$.
This enables us to assess model effectiveness across various operational scenarios.
Additionally, to understand the contribution of each feature to the model's predictions, we employed SHapley Additive exPlanations (SHAP) values \cite{lundberg2017a}.

\begin{equation}
\text{mROC} = \frac{1}{M} \sum_{c=1600}^{1900} \text{AUC-ROC}_c
\label{eq:mroc}
\end{equation}

\begin{equation}
\text{mPR} = \frac{1}{M} \sum_{c=1600}^{1900} \text{AUC-PR}_c
\label{eq:mpr}
\end{equation}

\subsection{Analysis of Trichome Density in Relation to Fertilizer Stress}
\subsubsection{Research objectives and scope}
This research primarily aims to develop practical agricultural tools for farmers to address cultivation challenges by investigating the relationship between fertilizer stress and trichome density in tomato plants.
Our approach leverages reverse inference, utilizing a known natural phenomenon where leaf trichome density changes in response to levels of fertilizer stress.
While our experimental design primarily focuses on creating practical tools for farmers to assess fertilizer stress through observed trichome density patterns, it also provides an opportunity to contribute, as a secondary outcome, to the broader scientific understanding of this biological response in tomato plants.

\subsubsection{Analysis methodology}
To quantitatively investigate the relationship between fertilizer stress and trichome density, we employed a machine learning regression approach as a forward inference step, addressing the interests of researchers studying the natural phenomenon itself, rather than its reverse inference application for farmers. 
The analysis was designed to investigate how much of the variation in leaflet trichome density under fertilizer stress conditions can be explained by fertilizer stress indicators.

Our model's prediction was built upon three key features.
We used nitrate ion concentration as the primary predictor, while image resolution and exposure time served as moderator variables for image acquisition parameters.
We implemented the LightGBM algorithm with default hyperparameters and configured it with an L2 loss objective function, 100 boosting rounds, and a learning rate of 0.1.

The model's performance and generalizability were rigorously evaluated via LOOCV, where each unique leaflet iteratively served as the test set.
We applied standard scaling to both the feature variables and the target variable before training.
Each LOOCV fold involved training a LightGBM model on the scaled training data to predict the NND, followed by inverse transformation to the original scale for evaluation.
Model performance was assessed via multiple metrics, including the root mean square error ($\mathrm{RMSE}$), coefficient of determination ($R_2$), and Pearson correlation coefficient ($r$).

\section{Results and Discussion}
\label{sec:results and discussion}
\subsection{Leaf Nitrate Concentration as a Lagging Indicator of Fertilizer Stress}
\begin{figure}[t!]
  \centering
  \includegraphics[width=0.5\columnwidth, keepaspectratio=true]{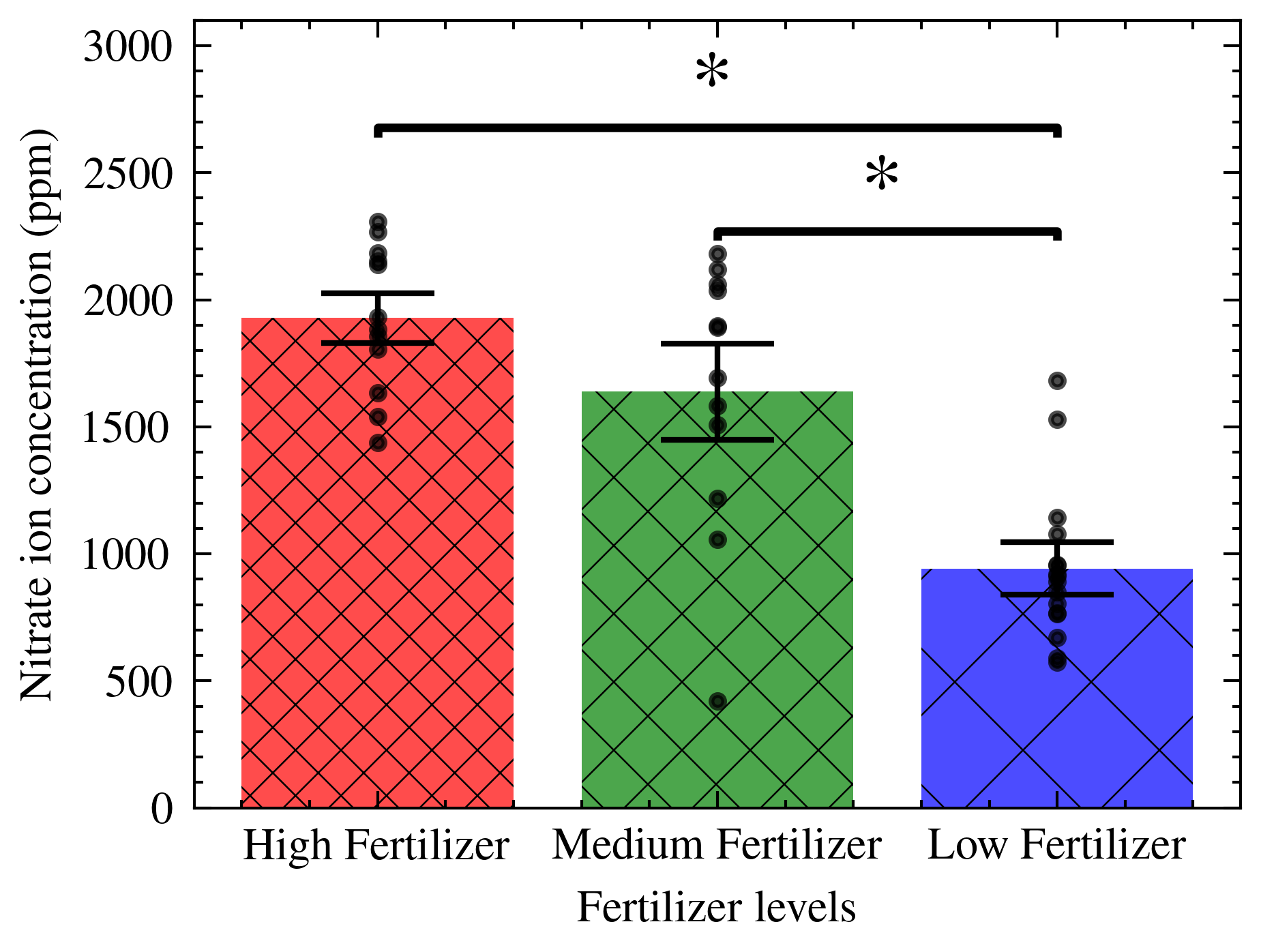}
  \caption{\textbf{Distribution of nitrate concentrations in compound leaves at different fertilizer levels.}}
  \label{fig10}
\end{figure}

The distribution of nitrate ion concentrations in compound leaves, measured 15 cm from the apical meristem across three fertilizer treatment levels (high, medium, and low), is presented in Figure \ref{fig10}.
Notably, higher fertilizer application rates corresponded to elevated nitrate ion concentrations.
Statistical analysis via the Kruskal-Wallis test revealed significant differences among the three groups ($p < 0.05$).
Pairwise comparisons via Mann-Whitney U tests with Bonferroni correction revealed significant differences between the high-low and medium-low groups ($\text{adjusted}\,p < 0.05$), whereas the high-medium comparison revealed no significant difference ($\text{adjusted}\,p = 0.52$).

The absence of a significant difference between the high- and medium-intensity fertilizer treatments suggests a diminishing returns effect.
These findings indicate that while the maximum yield is greater in the high-intensity fertilizer treatment than in the low-intensity treatment, additional fertilizer inputs beyond the medium level yield increasingly smaller gains.
While these results indicate that compound leaf nitrate ion concentration can serve as a reliable indicator of fertilizer stress in tomatoes, particularly for distinguishing between low and high fertilizer levels, the relationship shows notable variability.
As illustrated in Figure \ref{fig10}, the overlapping distributions between the high and medium treatments highlight this variability.

These findings emphasize the importance of directly measuring plant nitrate uptake rather than relying solely on fertilizer application rates to accurately evaluate fertilizer stress.
Direct measurement of nutrient uptake provides more comprehensive insight into nutrient availability by accounting for various factors beyond fertilizer application that may influence nutrient absorption.

\subsection{Leaf Nitrate Concentration as a Leading Indicator of Fruit Yield}
\begin{figure}[t!]
  \centering
  \includegraphics[width=0.5\columnwidth, keepaspectratio=true]{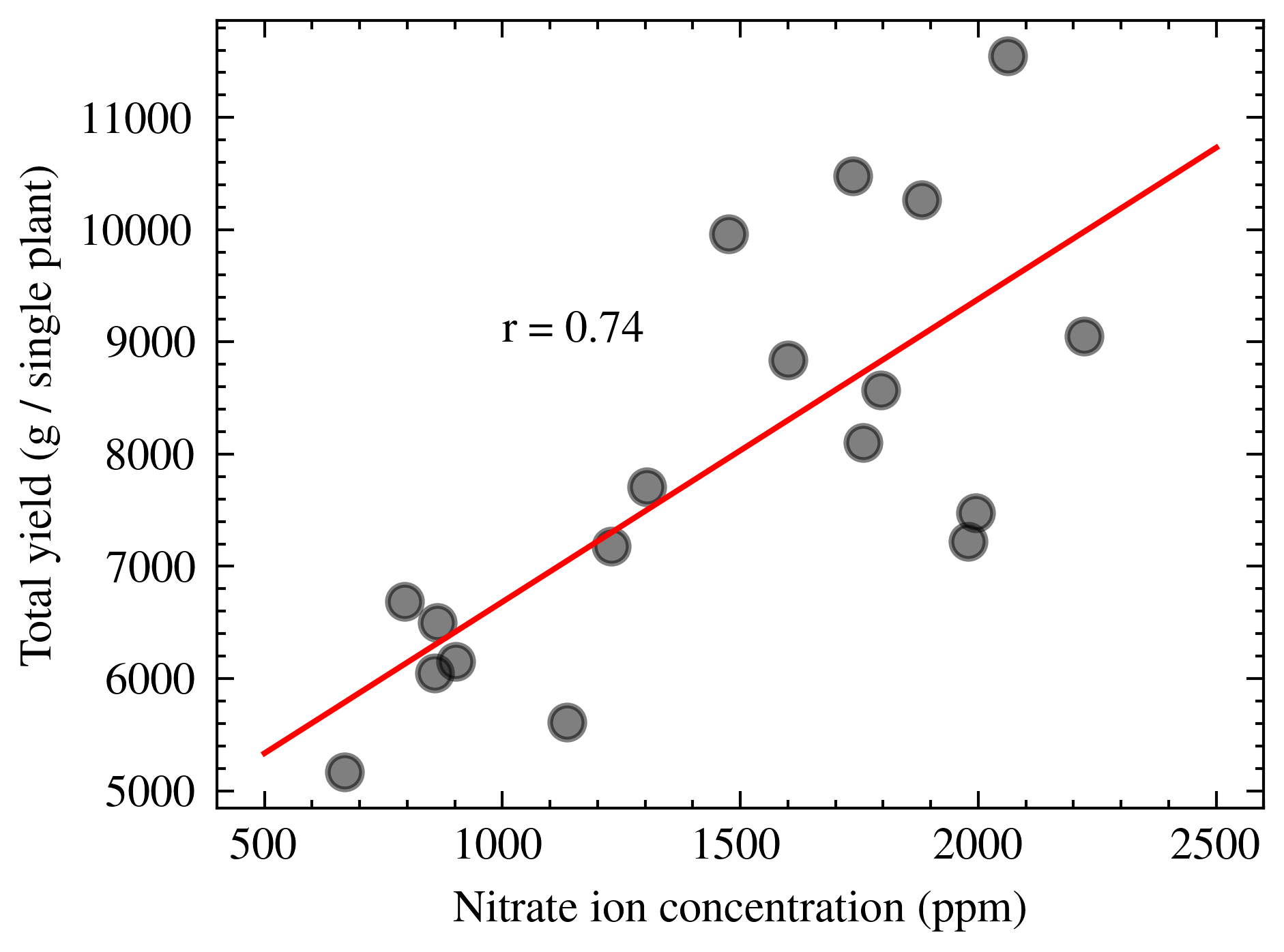}
  \caption{\textbf{Relationship between the nitrate ion concentration in compound leaves and yield.}}
  \label{fig11}
\end{figure}

Regarding the relationship between fruit yield and compound leaf nitrate concentration, the results demonstrated a positive correlation, as shown in Figure \ref{fig11}, indicating that leaf nitrate ion concentration can serve as a yield indicator.
While crop yield generally increases with fertilizer absorption, excessive application beyond a critical threshold can reduce yields \cite{andersen1999relationships,locascio1997nitrogen,coltman1988yields}.
Our experimental results suggest that fertilizer application remained below this critical threshold, indicating that the tomatoes experienced stress from insufficient fertilization.

\begin{figure}[t!]
  \centering
  \includegraphics[width=0.5\columnwidth, keepaspectratio=true]{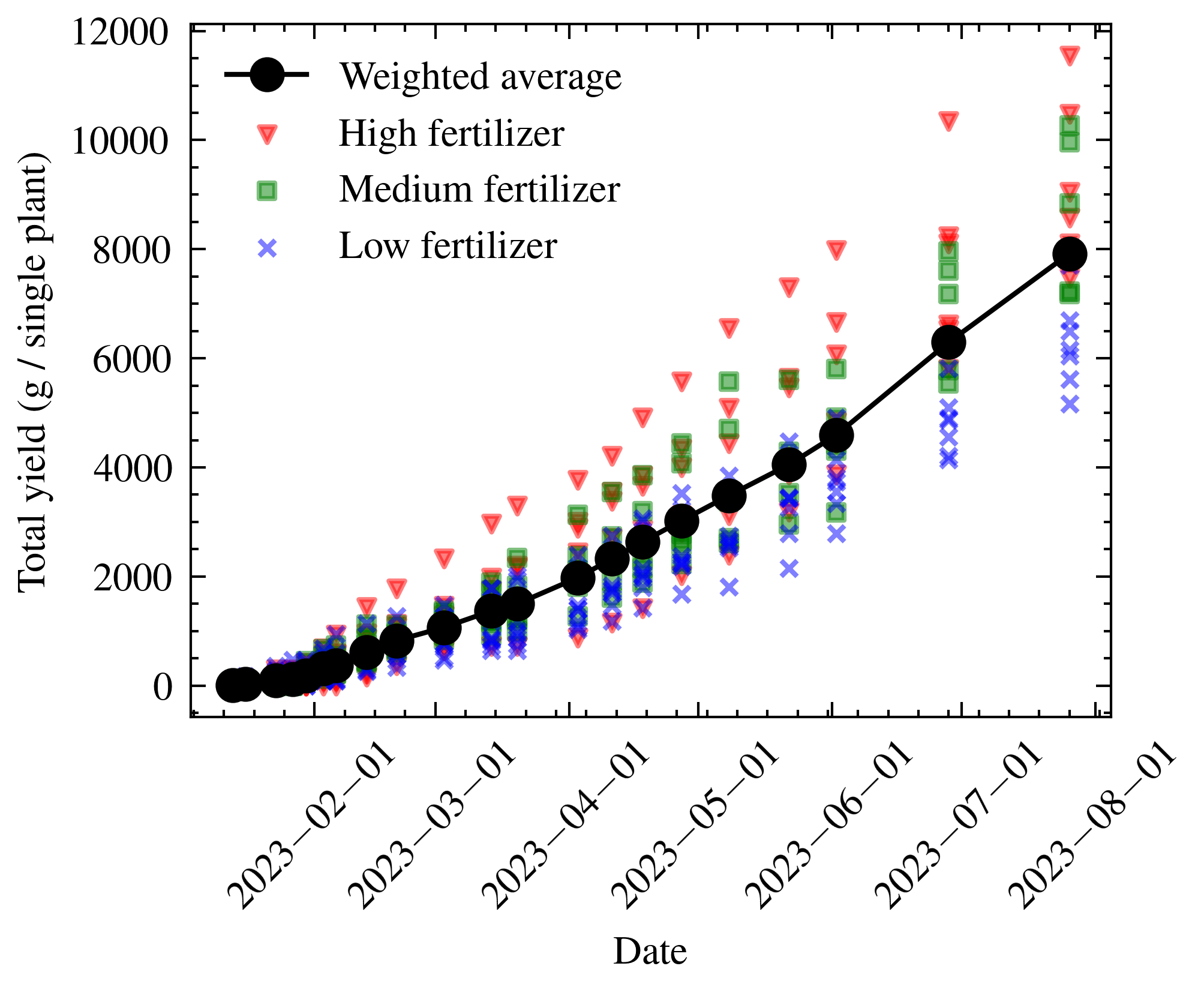}
  \caption{\textbf{Cumulative fruit yield in the experimental setup.}}
  \label{fig12}
\end{figure}

To evaluate yield potential distinctions, we employed a nitrate concentration threshold of 1,783 ppm, representing the third quartile of measured leaf concentrations.
The experimental setup's cumulative yield, presented in Figure \ref{fig12}, was calculated as the weighted average of individual intervention trial plots within the experimental environment, resulting in a final fruit production capacity of \SI{39.12}{\kilogram\per\square\meter} based on planting density.
This yield compares favorably with previously reported commercial tomato yields \cite{maureira2022evaluating}, suggesting that our experimental conditions, including temperature, humidity, light, water availability, oxygen levels, and carbon dioxide concentrations, effectively simulated commercial production environments.

\subsection{Handling of Multicollinearity and Variable Interaction in Data Analysis}
Our initial statistical analysis revealed significant multicollinearity among the explanatory variables (Farrar-Glauber test, $p < 0.001$) \cite{farrar1967multicollinearity}, indicating a statistically significant linear dependence among them.
An examination of VIFs showed that ISO speed ratings presented an exceptionally high VIF of 47.80, far exceeding the conventional threshold of 10 (Table \ref{table:vif_scores}).
This multicollinearity arose from the camera's automatic exposure (AE) function responding to image brightness, which was influenced by light reflection from trichomes that appeared as white points.
As trichome density varied with fertilizer stress, the resulting changes in image brightness triggered automatic adjustments in ISO sensitivity.
This mechanism positioned ISO sensitivity as a mediator variable, not through direct measurement but rather by reflecting the camera's response to image characteristics, which created an unintended correlation with both the variable of interest (trichome density) and the target variable (fertilizer stress).

\begin{table}[!t]
\renewcommand{\arraystretch}{1.3}
\caption{Examination of Multicollinearity through the VIF}
\label{table:vif_scores}
\centering
\begin{tabular}{ll}
\toprule
\textbf{Feature} & \textbf{VIF} \\
\midrule
Nearest Neighbor Distance & 43.60 \\
Resolution & 11.03 \\
Exposure Time & 4.98 \\
ISO Speed Ratings & 47.80 \\
\bottomrule
\end{tabular}
\end{table}

Although multicollinearity may not significantly impact the model's predictive performance during cross-validation, it can substantially affect the interpretation of individual variable importance, leading to unstable and unreliable coefficient estimates.
Given our study's focus on both generalization performance and model interpretability, we excluded ISO speed ratings from subsequent analyses.
After this exclusion, no remaining variables had VIF values exceeding 10 (Table \ref{table:vif_scores_without_iso}), resolving the multicollinearity concerns.

\begin{table}[!t]
\renewcommand{\arraystretch}{1.3}
\caption{VIF Value after Dimensionality Reduction}
\label{table:vif_scores_without_iso}
\centering
\begin{tabular}{ll}
\toprule
\textbf{Feature} & \textbf{VIF} \\
\midrule
Nearest Neighbor Distance & 8.63 \\
Resolution & 6.49 \\
Exposure Time & 3.10 \\
\bottomrule
\end{tabular}
\end{table}

\begin{figure}[t!]
  \centering
  \includegraphics[width=0.5\columnwidth, keepaspectratio=true]{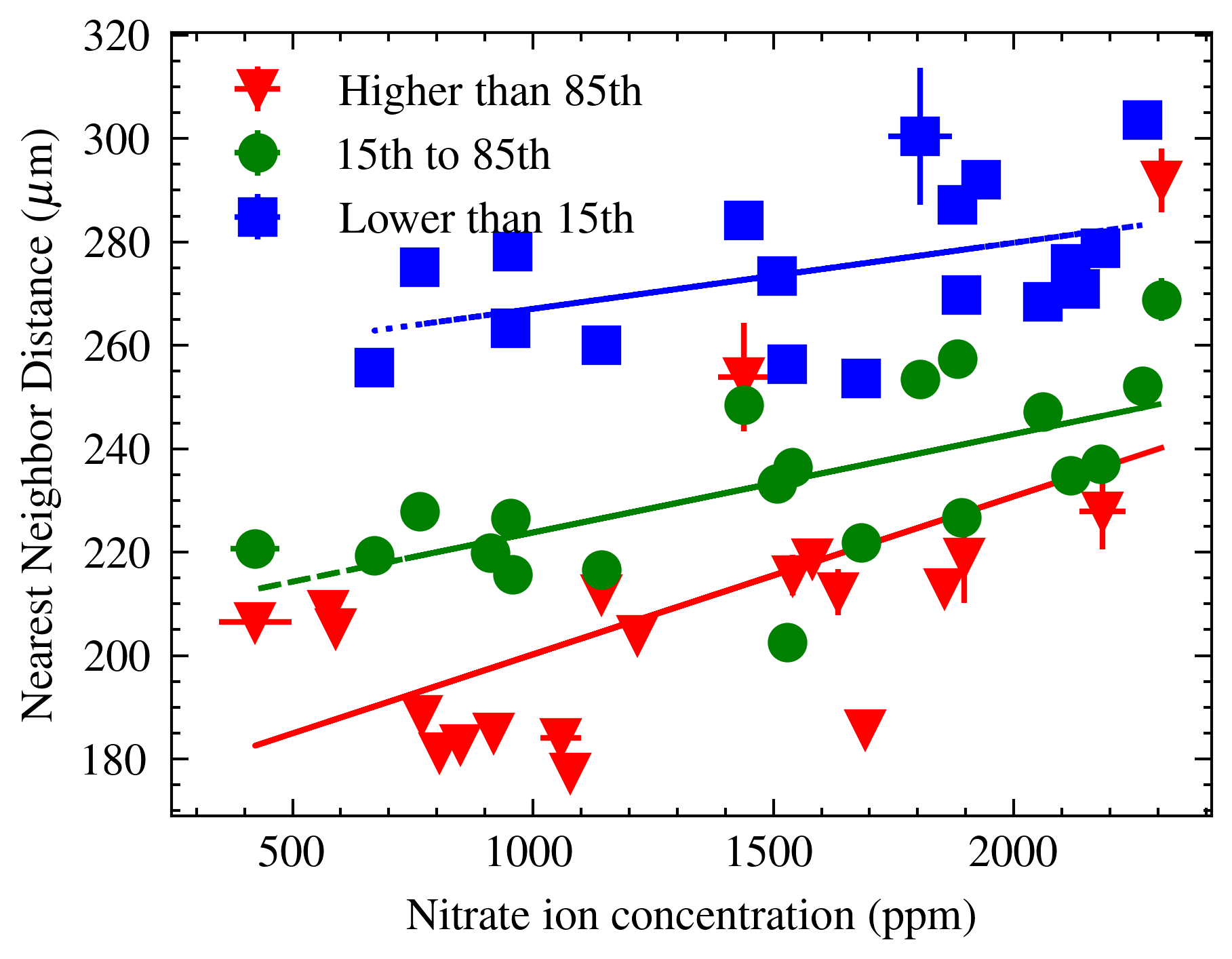}
  \caption{\textbf{Effect of resolution on the relationship between trichome density and nitrate concentration in leaves.}}
  \label{fig13}
\end{figure}

The resolution value, however, demonstrated interactions and intercept shifts in the relationship between trichome density and nitrate concentration, with varying coefficients and intercepts in the regression equation, as shown in Table \ref{table:resolution_percentile} and Figure \ref{fig13}.
Resolution variations can arise from multiple factors, including camera performance and the distance between the camera sensor and measurement paper.
In this study, as all images were captured via the same camera device, resolution primarily served as an indicator of the camera-to-paper distance.
To avoid omitted variable bias, which could lead to a misinterpretation of the effects of one variable as those of another, we included resolution as a key variable in the model.
As shown in Table \ref{table:resolution_percentile}, higher resolution values were associated with larger coefficients in our regression analysis, suggesting that optimal trichome detection in our computer vision application requires close camera proximity and high-resolution sensors.
Based on these analyses, our final model included the NND, resolution, and exposure time as predictor variables, each of which had independent effects on the response variable.

\begin{table}[!t]
\renewcommand{\arraystretch}{1.3}
\caption{Interaction when Stratified Based on Resolution}
\label{table:resolution_percentile}
\centering
\begin{tabular}{lcc}
\toprule
\textbf{Percentile} & \textbf{Resolution (pixel)} & \textbf{Equation} \\
\midrule
Higher than 85th & $>$ $1.08 \times 10^{6}$ & $y = 3.06 \times 10^{-2}x + 170$ \\
15th to 85th & $1.08 \times 10^{6} - 5.56 \times 10^{5}$ & $y = 1.90 \times 10^{-2}x + 205$ \\
Lower than 15th & $<$ $5.56 \times 10^{5}$ & $y = 1.28 \times 10^{-2}x + 254$  \\
\bottomrule
\end{tabular}
\end{table}

\subsection{Performance of the Proposed Method}
\begin{figure}[t!]
  \centering
  \includegraphics[width=0.5\columnwidth, keepaspectratio=true]{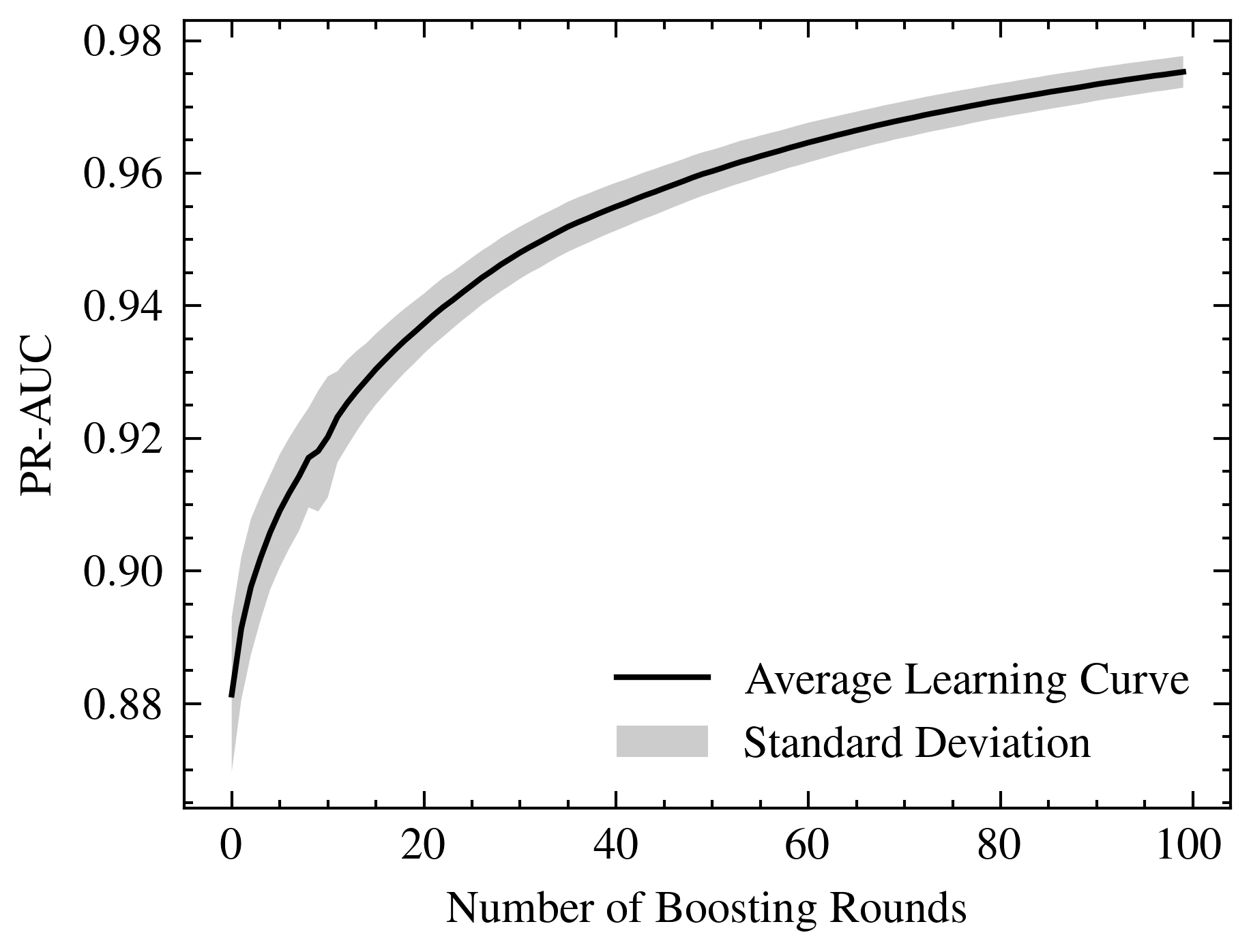}
  \caption{\textbf{Learning curve of the LightGBM model.}}
  \label{fig14}
\end{figure}

Our LightGBM model demonstrated consistent improvement in performance over 100 boosting rounds via default parameters, as evidenced by the gradual increase in AUC-PR values for training sets shown in Figure \ref{fig14}, indicating that the model was adequately trained.
For comprehensive evaluation, we established a representative scenario using a nitrate concentration threshold of 1,783 ppm with 25 images captured from a single compound leaf.
Under this scenario, the model obtained notable performance metrics with an mROC of 0.64 and an mPR of 0.82, despite variations in measurement data due to different optical conditions, enabling accurate assessment of additional fertilization needs.
Table \ref{table:predictive_performance} presents the model's predictive performance across different nitrate concentration thresholds, showing well-balanced precision and recall values.

\begin{table}[!t]
\renewcommand{\arraystretch}{1.3}
\caption{Predictive Performance in Terms of the Need for Additional Fertilizer using LOOCV at Various Nitrate Ion (ppm) Thresholds when Using the 25 Image Features}
\label{table:predictive_performance}
\centering
\begin{tabular}{lccc}
\toprule
\textbf{Evaluation Metrics} & \textbf{1600} & \textbf{1750} & \textbf{1900} \\
\midrule
Precision & 0.68 & 0.76 & 0.79 \\
Recall & 0.66 & 0.76 & 0.79 \\
F1 Score & 0.67 & 0.76 & 0.79 \\
PR AUC & 0.76 & 0.84 & 0.87 \\
ROC AUC & 0.64 & 0.68 & 0.58 \\
\bottomrule
\end{tabular}
\end{table}

\begin{figure}[t!]
  \centering
  \includegraphics[width=0.5\columnwidth, keepaspectratio=true]{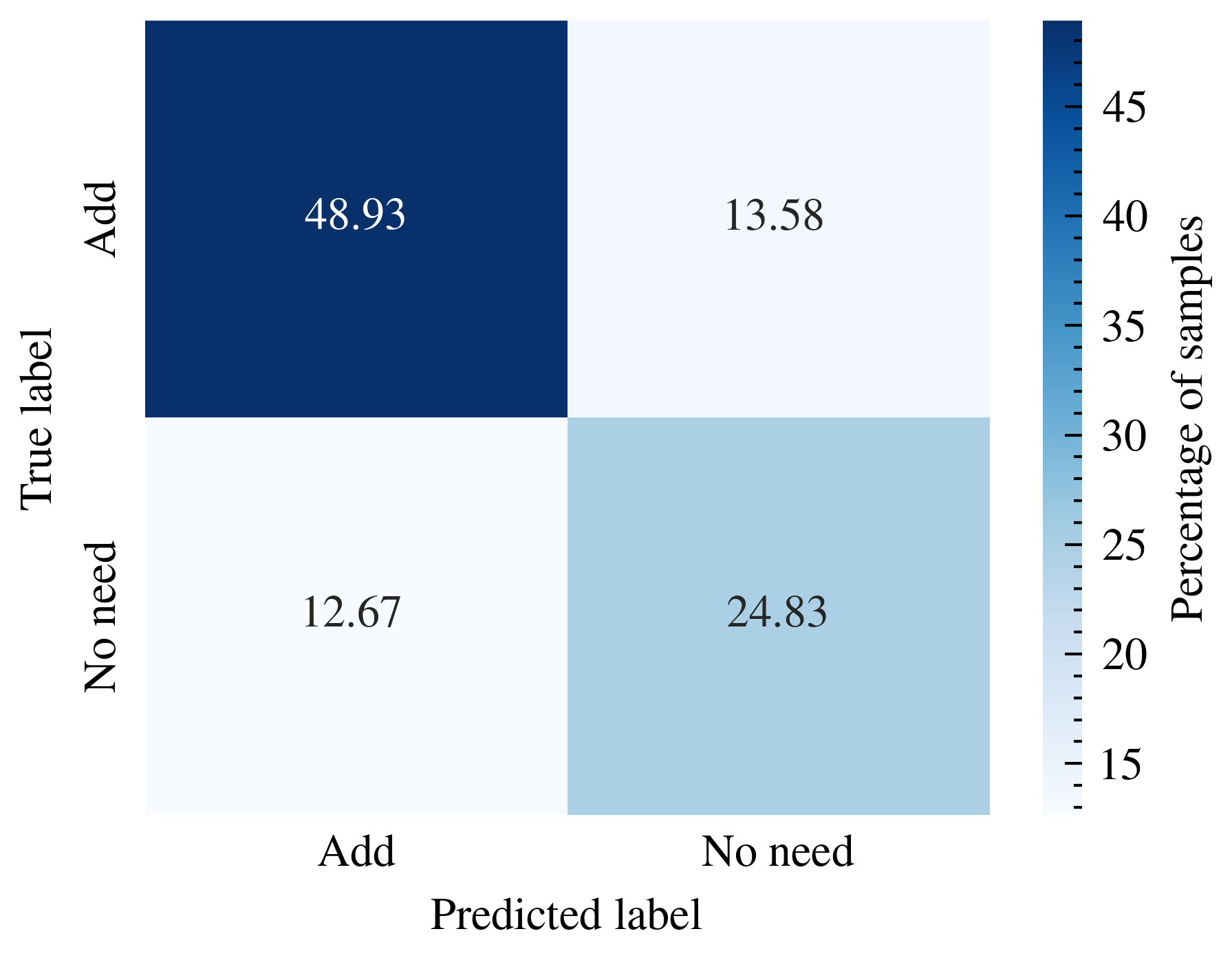}
  \caption{\textbf{Confusion matrix for classification outcomes during LOOCV in a representative parameter setting.}}
  \label{fig:confusion_matrix}
\end{figure}

\begin{figure}[t!]
  \centering
  \includegraphics[width=0.5\columnwidth, keepaspectratio=true]{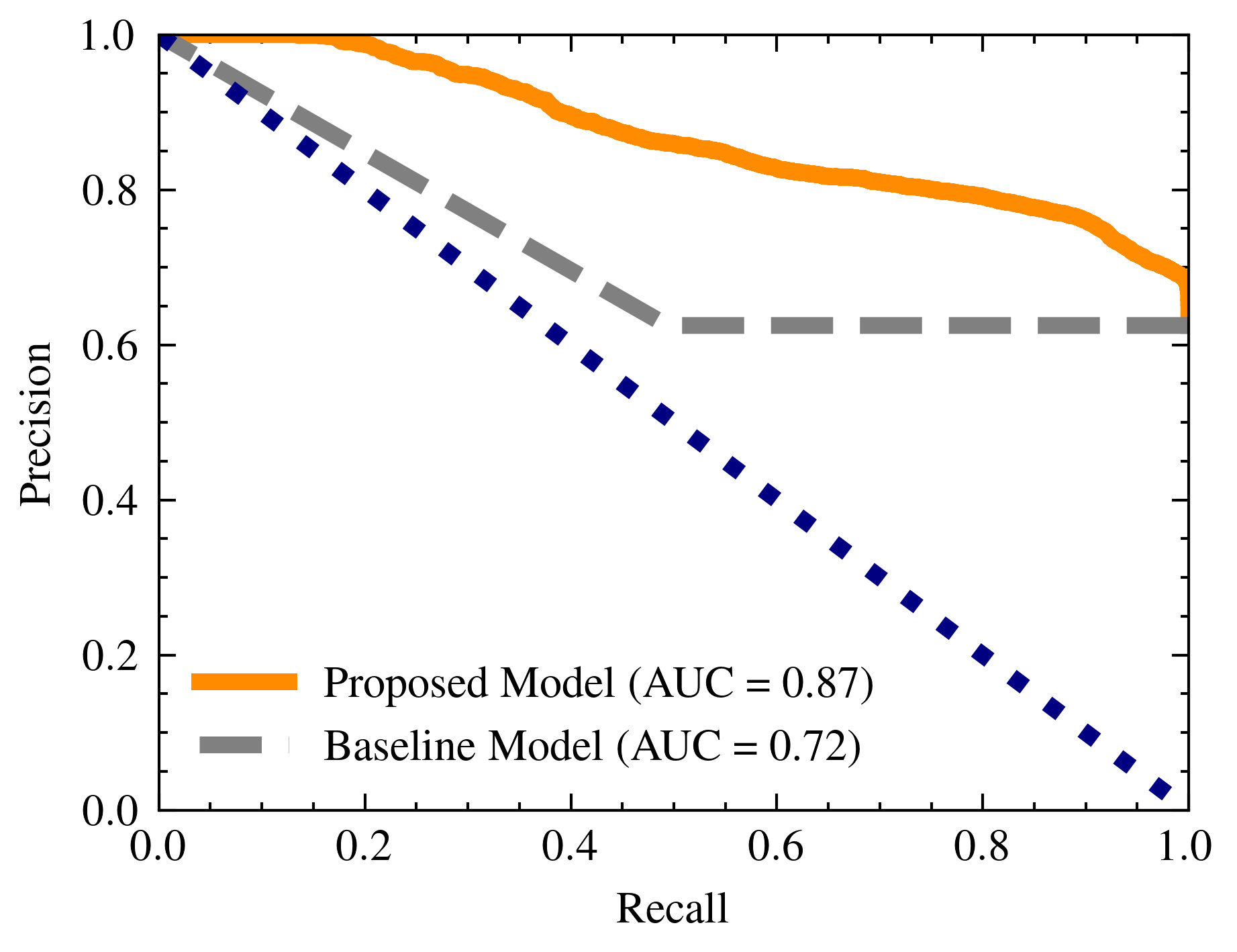}
  \caption{\textbf{Precision-recall (PR) curve analysis of the model during LOOCV in a representative parameter setting.}}
  \label{fig:pr_curve}
\end{figure}

\begin{figure}[t!]
  \centering
  \includegraphics[width=0.5\columnwidth, keepaspectratio=true]{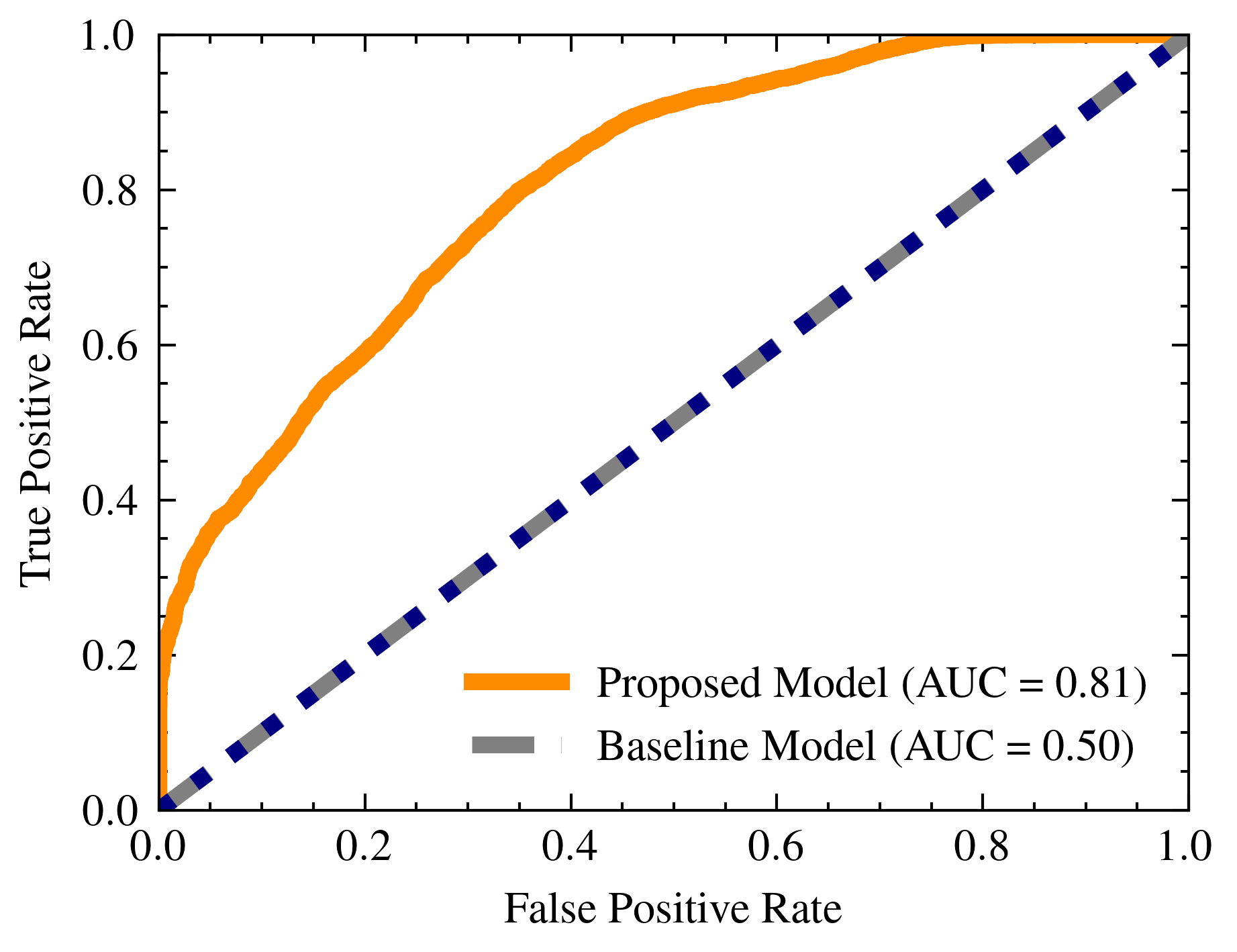}
  \caption{\textbf{Receiver operating characteristic (ROC) curve analysis of the model during LOOCV in a representative parameter setting.}}
  \label{fig:roc_curve}
\end{figure}

The model's balanced classification performance is evidenced by the confusion matrix in Figure \ref{fig:confusion_matrix}, which shows no substantial bias toward either false positives or false negatives.
This balanced performance is further supported by the F1 scores, AUC-PR values, and AUC-ROC values presented in Table \ref{table:predictive_performance} and visualized through the PR and ROC curves in Figures \ref{fig:pr_curve} and \ref{fig:roc_curve}, respectively.
Compared with a baseline majority class predictor model, our proposed model demonstrated superior performance across all evaluation metrics defined in Equations \ref{eq:7}-\ref{eq:11} (Table \ref{table:model_vs_dummy}).

\begin{table}[!t]
\renewcommand{\arraystretch}{1.3}
\caption{Comparison of Predictive Performance between the Proposed Model and the Baseline Model in a Representative Parameter Setting}
\label{table:model_vs_dummy}
\centering
\begin{tabular}{lccccc}
\toprule
\textbf{Model} & \textbf{Precision} & \textbf{Recall} & \textbf{F1 Score} & \textbf{ROC AUC} & \textbf{PR AUC} \\
\midrule
Proposed & 0.79 & 0.78 & 0.79 & 0.81 & 0.87 \\
Baseline & 0.62 & 0.49 & 0.55 & 0.50 & 0.72 \\
\bottomrule
\end{tabular}
\end{table}

\begin{figure}[t!]
  \centering
  \includegraphics[width=0.5\columnwidth, keepaspectratio=true]{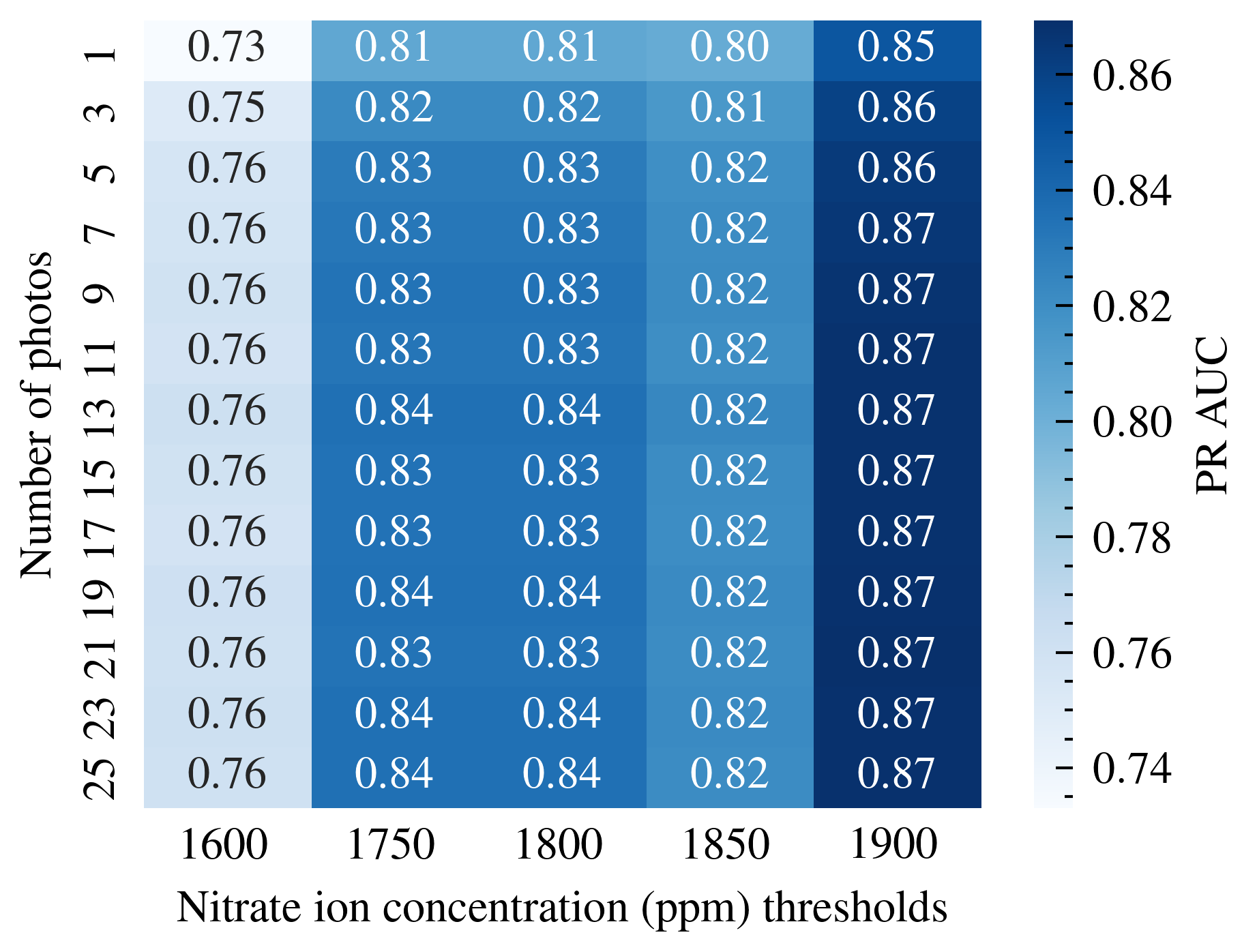}
  \caption{\textbf{Model prediction performance when various thresholds and numbers of images are set.}}
  \label{fig16}
\end{figure}

\begin{figure}[t!]
  \centering
  \includegraphics[width=0.5\columnwidth, keepaspectratio=true]{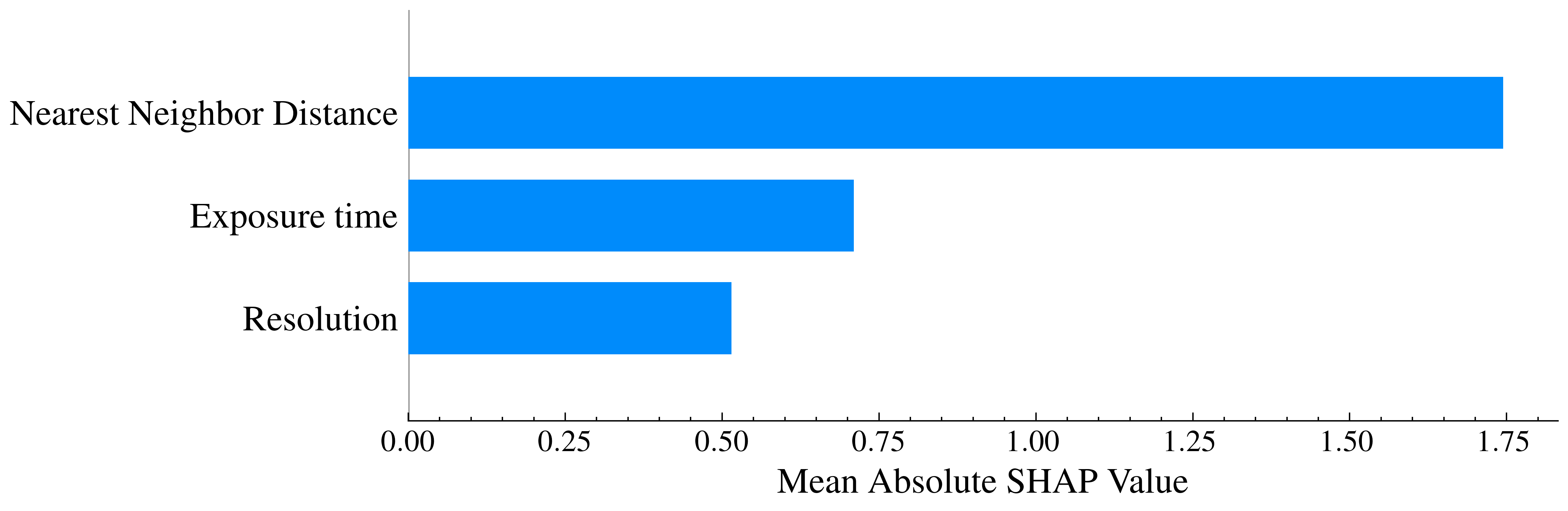}
  \caption{\textbf{Feature contribution during LOOCV with a representative model setting.}}
  \label{fig17}
\end{figure}

Analysis of the model's practical applicability revealed that even with a single input image, the model obtains a mean AUC-PR (mPR) of 0.80, with performance improving as the number of input images increases (Figure \ref{fig16}).
The feature contribution analysis using SHAP values (Figure \ref{fig17}) identified the NND as the most significant contributor to the model's predictions, showing the highest mean absolute SHAP value across the dataset.

The performance metrics demonstrate the practical viability of a smartphone-based trichome density measurement system. With just a smartphone camera, measurement sheets and analysis software, our method automates traditional microscope-based counting for the rapid assessment of tomato plant nutrient status.

\subsection{Effect of Fertilizer on Trichome Density}

\begin{figure}[t!]
  \centering
  \includegraphics[width=0.5\columnwidth, keepaspectratio=true]{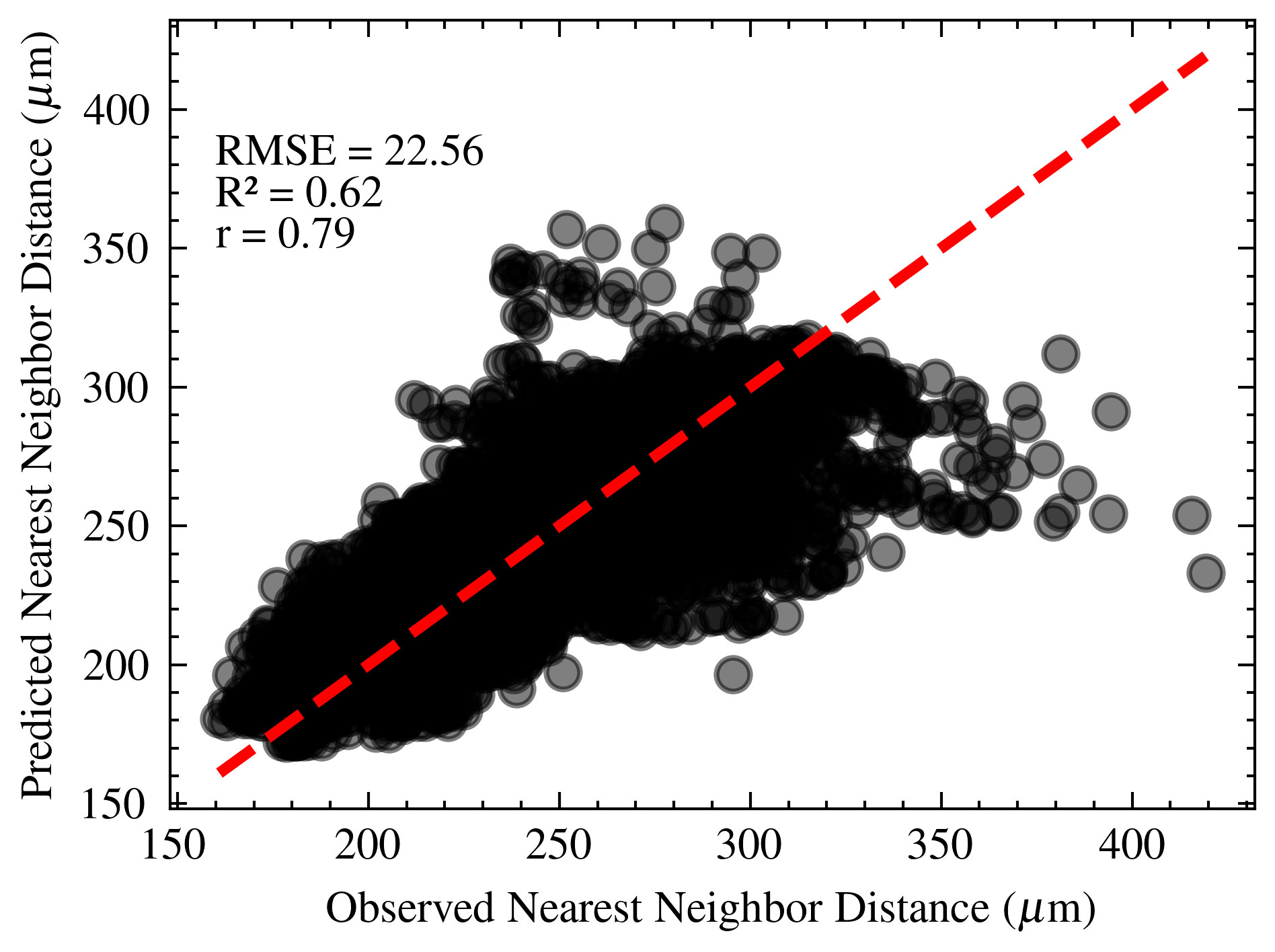}
  \caption{\textbf{Modeling of the NND via nitrate concentration, resolution, and exposure time.}}
  \label{fig18}
\end{figure}

Our quantitative evaluation of the relationship between tomato leaf trichome density and nitrogen nutritional status employed a prediction model using the NND as a measure of trichome density.
As shown in Figure \ref{fig18}, the model incorporated nitrate ion concentration as an explanatory variable, while image resolution and exposure time served as moderating variables.
This combination of variables explained 62.48\% of the variation in trichome density ( $R^2$ = 0.63) observed in leaflets from our fertilizer intervention experiments.
The remaining unexplained variation (37.52\%) suggests opportunities for future research to investigate additional influencing factors.

Importantly, while our study provides evidence supporting the relationship between fertilizer stress and trichome density, our approach was primarily data driven rather than being focused on rigorous hypothesis testing for interpreting natural phenomena in complex systems.
Unlike traditional hypothesis-based research, which typically begins with specific hypotheses and theories for deductive testing \cite{glass2008brief}, our study generated hypotheses and models inductively from the patterns and relationships within the dataset, guided by basic hypotheses, as illustrated in Figure \ref{fig3}.
While potentially introducing inherent biases when extended to the interpretation of underlying biological phenomena, this methodology provides adequate evidence for technological validation in the context of kit development.

These potential methodological biases are not considered fatal in our context because previous hypothesis-based research has already established the quantitative relationship between tomato leaf trichome density and nitrogen nutritional status \cite{hoffland2000nitrogen}. Additionally, our actual task was to capture established general trends for practical applications rather than identifying subtle biological differences in complex plant systems where hypothesis-based research would be prioritized.
Our findings support the influence of fertilizer stress on trichome density in young leaves, although further investigations could enhance model performance and our understanding of this relationship.

\subsection{Method Performance and Limitations}

Tomatoes pose a challenge in detecting fertilizer stress through noncontact optical measurements of young leaf chlorophyll.
This challenge is due to the nonlinear relationship between optical measurements and leaf nitrogen content, which distinguishes tomatoes from other crops \cite{padilla2018proximal,gianquinto2011a}.
To address this problem, our study focused on trichomes, which are densely packed hair-like structures on leaf surfaces that have been less explored in previous research.
We investigated the feasibility of detecting fertilizer stress by developing a device that utilizes trichome density as a primary predictor.
Our findings suggest the possibility of overcoming the limitations of conventional noncontact optical methods in assessing tomato fertilizer conditions, although our diagnostic kits have certain performance limitations.

To validate this approach, we conducted experiments using a well-designed setup that simulated different levels of fertilizer stress common in agricultural production.
Our cost-effective hydroponic cultivation system with plastic buckets achieved yields comparable to those of commercial cultivation, suggesting that agricultural conditions were effectively simulated.
However, we acknowledge that not all farmers may be able to establish optimal growing conditions due to equipment costs.
As a result, environmental stresses beyond fertilizer deficiency could decrease our method's predictive performance.

\subsection{Practical Implications}

Traditional nutrient status assessment through visual inspection of leaf color, size, texture, and overall plant health heavily relies on individual farmers' experience and motivation, leading to considerable variations and standardization challenges.
Even when specific machines or systems are utilized for nutrient status assessment, the combinations of phenotypic indicators and sensor types create endless possibilities for evaluating utility across multiple dimensions, including the complexity of data processing, the prerequisites for measurement, operational complexity, time requirements, and detection latency from stress onset.
This variability creates situations where certain measurement methods work well under specific conditions, making it difficult to evaluate the effectiveness of diagnostic kits through predictive accuracy comparisons alone.
More comprehensive validation using lagging indicators, particularly financial metrics, is necessary to assess practical applicability in real-world settings.
While various commercial nutrient diagnostic kits exist, their high costs and application challenges with delicate tissues such as young tomato leaves, which are typically sensitive to nutrient deficiencies and crucial for plant growth and development, represent one aspect of the diverse factors affecting their adoption in practical farming settings.
In response, we propose a novel, simple, and cost-effective method specifically designed for assessing nutrient status in these delicate tissues.

\subsection{Future Research}

The potential applications of our trichome-based method extend beyond tomatoes to other agriculturally important species with abundant trichomes, including watermelon (\textit{Citrullus lanatus}), eggplant (\textit{Solanum melongena}), tobacco (\textit{Nicotiana tabacum}), soybean (\textit{Glycine max}), cannabis (\textit{Cannabis sativa}), and potato (\textit{Solanum tuberosum}).
This broader applicability could improve real-time nutrient management across various crops, potentially optimizing fertilizer application, reducing waste, and lowering environmental impact.
These improvements would ultimately contribute to more sustainable agricultural practices.
Further research is needed to validate the effectiveness of this method across different species and to investigate its potential integration with other precision agriculture technologies for comprehensive nutrient management strategies.

\section{Conclusions}
\label{sec:conclusions}
This study focused on developing a novel method for detecting fertilizer stress in tomato plants by measuring the density of trichomes on the leaves of these plants, including the setup and evaluation of necessary experimental conditions.
Our developed diagnostic kit enables noninvasive and early detection of fertilizer stress in plants without requiring customized sensing devices.
This universal approach provides accurate nutrient status assessment at a lower cost than traditional devices.
The findings present a novel approach for designing diagnostic devices for plant fertilizer stress detection by utilizing the leaf surface structure, offering potential advancements in agricultural practice.

\appendix \def\thesection{\Alph{section}}\section*{Appendix}
\section{Simulation Demonstrating the Robustness of the Nearest Neighbor Distance}
This appendix presents a simulation demonstrating the robustness of the nearest neighbor distance (NND) as a density metric compared with the conventional point count per unit area.
The simulation involves generating a set of points within a defined area and simulating damage by removing a portion of these points to compare the changes in both the NND and point count measurements.
To mimic the random spatial distribution of trichomes on a leaf surface, we employed a Poisson point process for point generation, as this process effectively models random events occurring independently in a given space or time.
This approach is particularly suitable for simulating trichome distribution, as trichomes are expected to emerge randomly on the leaf surface.
The detailed steps of this simulation are outlined in Algorithm \ref{alg:trichome_distribution_simulation}.

\begin{algorithm}
  \caption{Trichome distribution simulation}
  \label{alg:trichome_distribution_simulation}
  \begin{algorithmic}[1]
  \Require $\lambda$: Expected number of events (trichomes)
  \Require $A$: Total area of the region
  \Require $S$: Random seed value
  \Ensure $P$: Set of points representing trichome locations
  \Ensure $P'$: Set of points after simulated damage
  \Function{SetSeed}{$S$}
  \Statex Set a random seed to $S$
  \EndFunction
  \Function{PoissonRandom}{$\lambda$}
  \Statex Generate random numbers from the Poisson distribution
  \Return number of points
  \EndFunction
  \Function{UniformRandom}{$a, b$}
  \Statex Generate random numbers from a uniform distribution
  \Return random number between $a$ and $b$
  \EndFunction
  \Function{GeneratePoints}{$N$}
  \Statex Generate $N$ random points in the area
  \Return set of points $P$
  \EndFunction
  \Function{SimulateDamage}{$P$}
  \Statex Remove points with $x > 500$
  \Return set of remaining points $P'$
  \EndFunction
  \Statex
  \Statex \textbf{Main Execution}
  \State \Call{SetSeed}{$S$}
  \State $N \gets \Call{PoissonRandom}{\lambda}$
  \State $P \gets \Call{GeneratePoints}{N}$
  \State $P' \gets \Call{SimulateDamage}{P}$
  \end{algorithmic}
\end{algorithm}

\begin{figure}[t!]
  \centering
  \includegraphics[width=0.5\columnwidth, keepaspectratio=true]{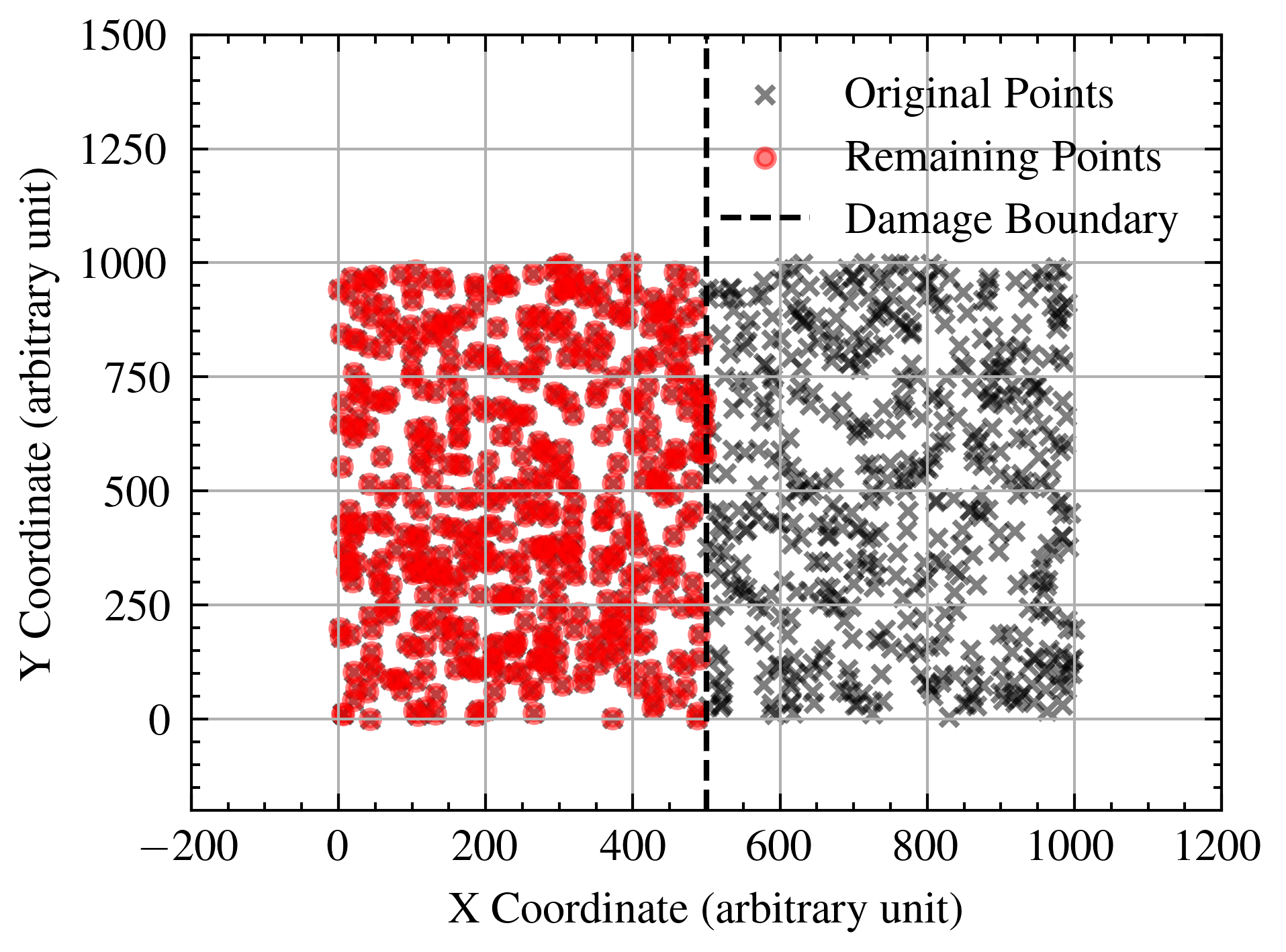}
  \caption{\textbf{Point distributions before and after simulated damage.}}
  \label{fig:point_distribution}
\end{figure}

\begin{figure}[t!]
  \centering
  \includegraphics[width=0.5\columnwidth, keepaspectratio=true]{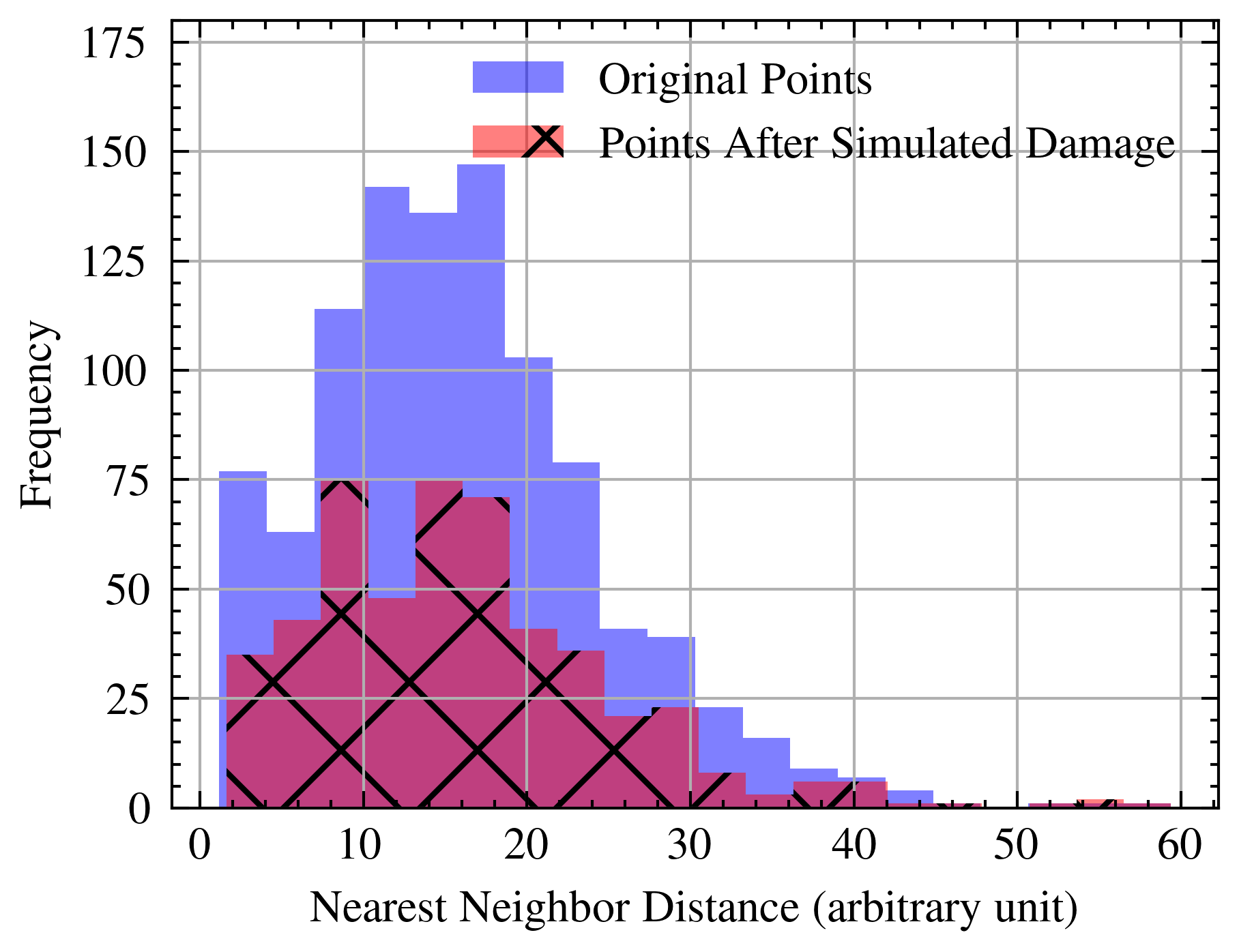}
  \caption{\textbf{Histogram of nearest neighbor distances.}}
  \label{fig:nnd_histogram}
\end{figure}

Figure \ref{fig:point_distribution} shows an example of the point distributions before and after simulated damage.
For this demonstration, we randomly generated points via the Poisson process within a 1000 $\times$ 1000 area.
Damage was simulated by removing points within the right half of the area, marked by the "damage boundary" line.
Figure \ref{fig:nnd_histogram} presents the histograms of NNDs before and after this simulated damage.

\begin{figure}[t!]
  \centering
  \includegraphics[width=0.5\columnwidth, keepaspectratio=true]{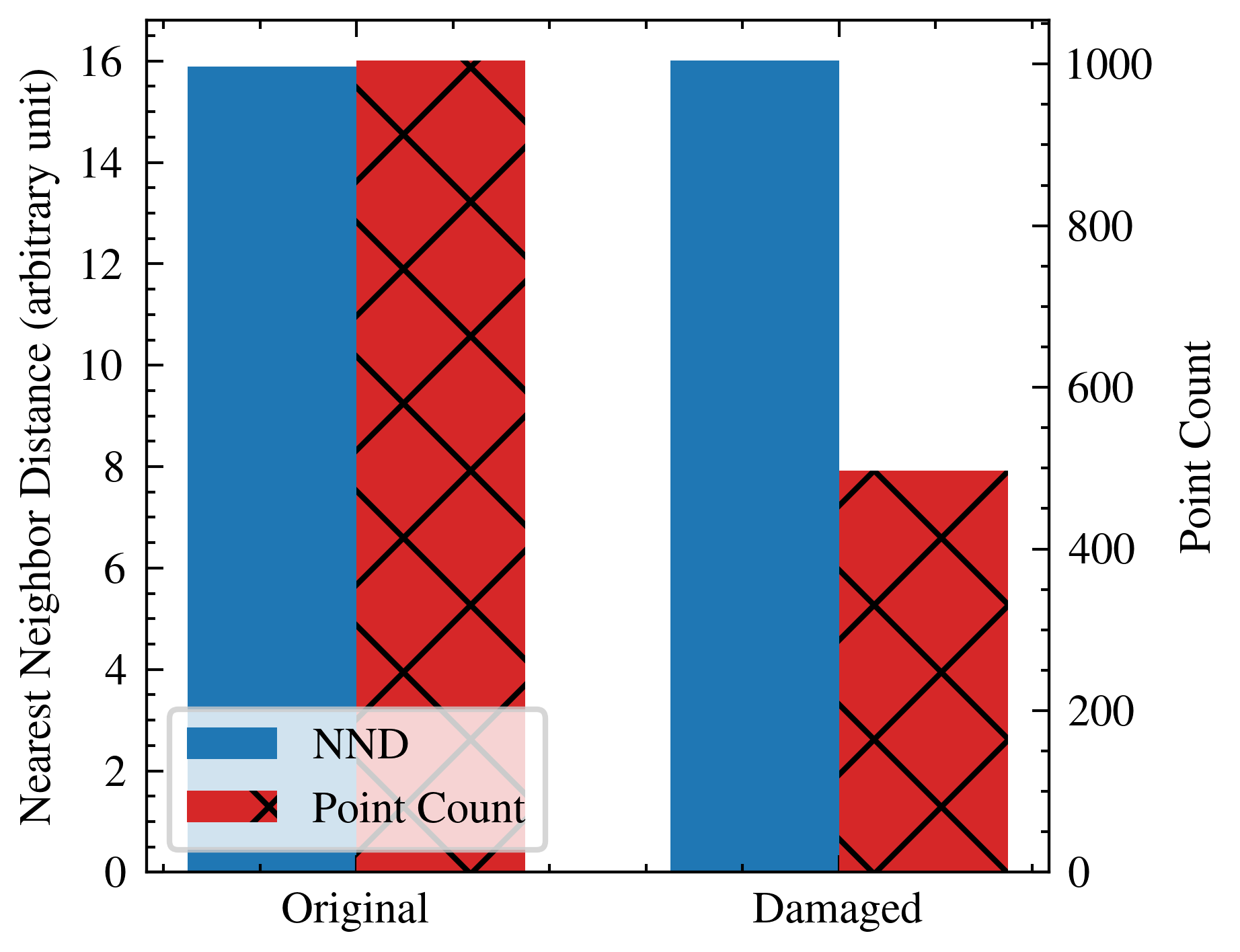}
  \caption{\textbf{Comparison of NND and point count after simulated damage.}}
  \label{fig:nnd_pointcount_comparison}
\end{figure}

\begin{figure}[t!]
  \centering
  \includegraphics[width=0.5\columnwidth, keepaspectratio=true]{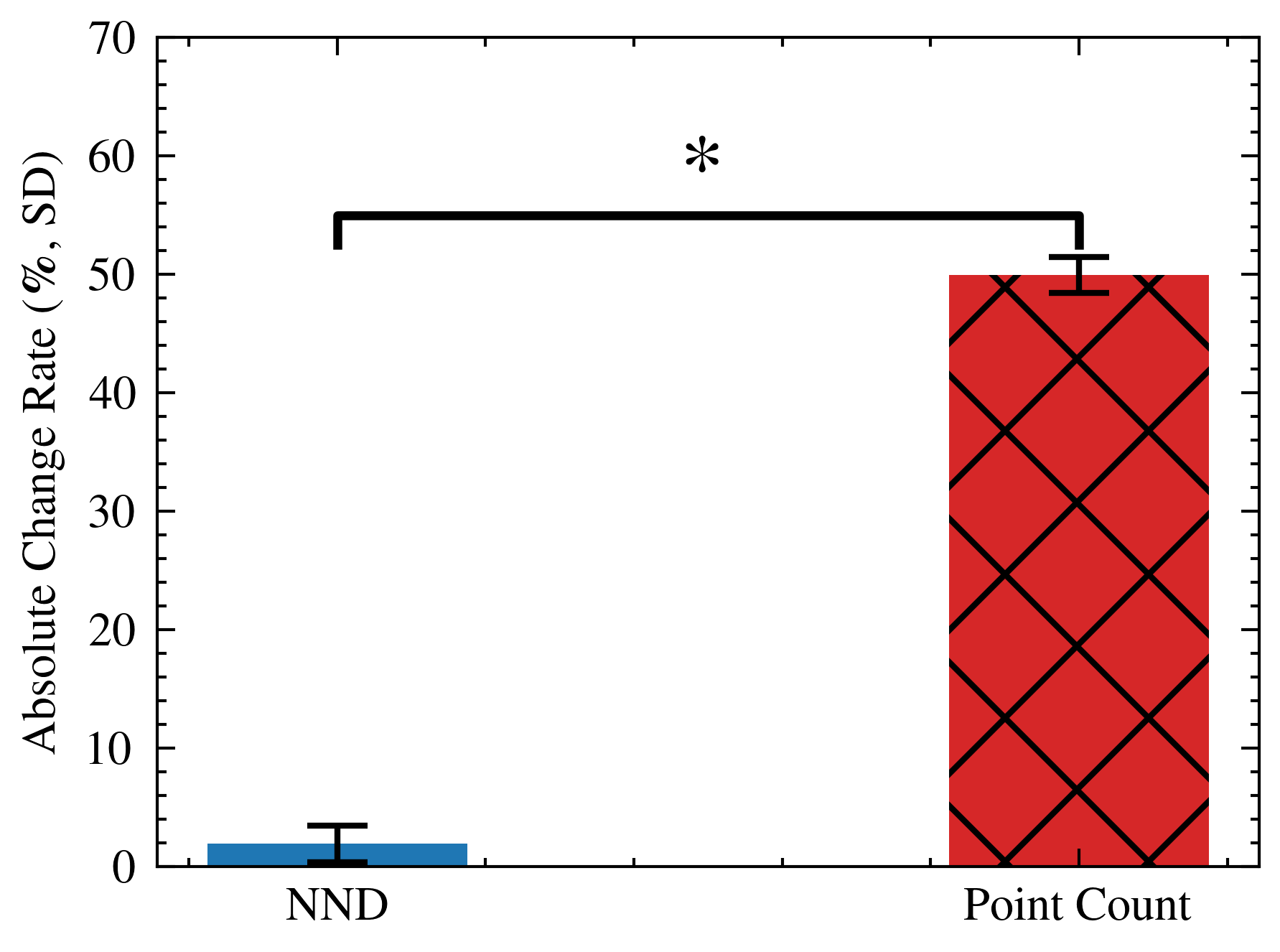}
  \caption{\textbf{Comparison of NND and point count absolute change rates after simulated damage.}}
  \label{fig:nnd_pointcount_change_rate}
\end{figure}

The simulation results demonstrate that the NND is more robust to point removal than the conventional density metric based on the point count per unit area.
Figure \ref{fig:nnd_pointcount_comparison} compares the NND and point count before and after simulated damage.
Although both metrics changed after damage, the change in the NND was smaller than that in the point count.
To validate this observation statistically, we performed a Wilcoxon signed-rank test to compare the absolute change rates of the NND and the point count after damage, as shown in Figure \ref{fig:nnd_pointcount_change_rate}.
The test revealed a statistically significant difference between the two metrics ($p < 0.001$), supporting the hypothesis that the NND is less affected by localized point loss than the point count per unit area.
This finding demonstrates the advantage of using the NND as a density metric for scenarios where localized point loss might occur, such as trichome detachment from leaf surfaces.

\section{Data analysis workflow}
Algorithm \ref{alg:fertilization_prediction} presents a comprehensive implementation of our tomato fertilization requirement prediction system.
The algorithm consists of four main functional components: label setting, class balance adjustment, model training, and evaluation.
The label setting function (SetLabels) processes the raw nitrate concentration data into binary classification labels based on the predefined threshold.
The SMOTE application function (ApplySMOTE) addresses class imbalance issues in the training data.
The model training function (TrainModel) handles the core prediction task, while the evaluation function (Evaluate) assesses model performance.
The main execution flow implements a leave-one-out cross-validation strategy, where data are iteratively partitioned by the compound leaf index.
For each iteration, the algorithm extracts test data for the current compound leaf while using the remaining data for training.
After applying SMOTE to balance the training data, the model is trained and evaluated, with the results accumulated for both the training and test sets.
The algorithm tracks the true and predicted values throughout the process and computes SHAP values.
Finally, it calculates comprehensive performance metrics across all iterations.

\begin{algorithm}
  \caption{Tomato fertilization requirement prediction}
  \label{alg:fertilization_prediction}
  \begin{algorithmic}[1]
  \Require $X$: Explanatory and moderator variables
  \Require $Y$: Target variables
  \Require $t$: Threshold for nitrate ion concentration
  \Require $n$: Measurement intensity
  \Require $M$: Metadata containing CompoundLeafIndex etc.
  \Ensure $metrics$: Performance metrics
  \Ensure $S$: SHAP values
  
  \Function{SetLabels}{$Y, t, M$}
  \State $L \gets \emptyset$
  \For{$i \gets 1$ to $|Y|$}
      \State $label \gets Y[i] < t ? 0 : 1$
      \State $M[i][\text{`ClassLabelIndex'}] \gets label$
      \State $L \gets L \cup \{label\}$
  \EndFor
  \State \Return $M, L$
  \EndFunction
  
  \Function{ApplySMOTE}{$X, Y, M$}
      \State Apply SMOTE to $X$ and $Y$ based on $M$
      \State \Return balanced $X, Y$
  \EndFunction
  
  \Function{TrainModel}{$X, Y$}
      \State Train model on $X, Y$
      \State $Y{pred} \gets \Call{MakePredictions}{model, X}$
      \State \Return $Y, Y{pred}, model$
  \EndFunction
  
  \Function{Evaluate}{$model, X, Y, n, M{test}$}
      \State MakePrediction
      \State \Return $Y, Y_{pred}$
  \EndFunction
  \Statex
  \Statex \textbf{Main Execution}
  \State $M, L \gets \Call{SetLabels}{Y, t, M}$
  \State $Y{train\_true},Y{train\_pred},Y{test\_true},Y{test\_pred} \gets \emptyset$
  \State $S \gets \emptyset$
  \State $I \gets \Call{ExtractCompoundLeafIndex}{M}$
  
  \For{each unique index $i$ in $I$}
      \State $X{test} \gets$ Select rows from $X$ where $I$ equals $i$
      \State $Y{test} \gets$ Select rows from $Y$ where $I$ equals $i$
      \State $M{test} \gets$ Select rows from $M$ where $I$ equals $i$
      \State $X{train} \gets$ All rows from $X$ except where $I$ equals $i$
      \State $Y{train} \gets$ All rows from $Y$ except where $I$ equals $i$
      \State $M{train} \gets$ All rows from $M$ except where $I$ equals $i$
      \State $X{train}, Y{train} \gets \Call{ApplySMOTE}{X{train}, Y{train}, M{train}}$
      \State $Y{true}, Y{pred}, model \gets \Call{TrainModel}{X{train}, Y{train}}$
      \State $Y{train\_true} \gets Y{train\_true} \cup Y{true}$
      \State $Y{train\_pred} \gets Y{train\_pred} \cup Y{pred}$
      \State $Y{test}, Y{pred} \gets \Call{Evaluate}{model, X{test}, Y{test}, n, M{test}}$
      \State $Y{test\_true} \gets Y{test\_true} \cup Y{test}$
      \State $Y{test\_pred} \gets Y{test\_pred} \cup Y{pred}$
      \State $S \gets S \cup \Call{ComputeSHAP}{}$
  \EndFor
  
  \State $metrics \gets \Call{CalculateMetrics}{}$
  
\end{algorithmic}
\end{algorithm}

\section{Comprehensive Model Performance across All Nitrate Ion Concentration Thresholds}

This section extends our model evaluation beyond the 1600-1900 ppm nitrate ion concentration range presented in Table \ref{table:predictive_performance} of the main text to encompass the entire range of observed concentrations in our study.
For a comprehensive assessment, we divided the full range of observed nitrate ion concentrations into 10 equal intervals, with each serving as a threshold for model evaluation.
This approach allows for a thorough assessment of model performance across all possible thresholds within our dataset.
Table \ref{table:predictive_performance_comparison} presents a detailed performance comparison between our proposed model and the baseline model, using LOOCV and features extracted from 25 images per compound leaf.
The proposed model's results are displayed as main values, with corresponding baseline model results shown in parentheses below, enabling direct performance comparison across all threshold intervals.

\begin{table}[!t]
\caption{Predictive Performance Comparison between Proposed and Baseline Models using LOOCV at Various Nitrate Ion Concentration Thresholds}
\label{table:predictive_performance_comparison}
\centering
\begin{tabular}{cccccc}
\toprule
\textbf{Threshold} & \textbf{Precision} & \textbf{Recall} & \textbf{F1 Score} & \textbf{PR AUC} & \textbf{ROC AUC} \\
\midrule
\multirow{2}{*}{600} & 0.15 & 0.04 & 0.06 & 0.13 & 0.51 \\
& (0.08) & (0.49) & (0.13) & (0.30) & (0.50) \\
\multirow{2}{*}{777} & 0.12 & 0.11 & 0.11 & 0.18 & 0.48 \\
& (0.15) & (0.49) & (0.23) & (0.36) & (0.50) \\
\multirow{2}{*}{955} & 0.39 & 0.42 & 0.40 & 0.49 & 0.57 \\
& (0.30) & (0.49) & (0.37) & (0.47) & (0.50) \\
\multirow{2}{*}{1133} & 0.60 & 0.66 & 0.63 & 0.69 & 0.70 \\
& (0.38) & (0.49) & (0.43) & (0.53) & (0.50) \\
\multirow{2}{*}{1311} & 0.61 & 0.66 & 0.63 & 0.71 & 0.67 \\
& (0.42) & (0.49) & (0.46) & (0.57) & (0.50) \\
\multirow{2}{*}{1488} & 0.61 & 0.61 & 0.61 & 0.70 & 0.65 \\
& (0.45) & (0.49) & (0.47) & (0.59) & (0.50) \\
\multirow{2}{*}{1666} & 0.71 & 0.68 & 0.69 & 0.79 & 0.65 \\
& (0.57) & (0.49) & (0.53) & (0.68) & (0.50) \\
\multirow{2}{*}{1844} & 0.74 & 0.72 & 0.73 & 0.82 & 0.63 \\
& (0.78) & (0.49) & (0.60) & (0.83) & (0.50) \\
\multirow{2}{*}{2200} & 0.95 & 1.00 & 0.97 & 0.97 & 0.50 \\
& (0.95) & (0.49) & (0.65) & (0.96) & (0.50) \\
\bottomrule
\end{tabular}
\end{table}

\section*{Acknowledgments}
The authors would like to express their sincere gratitude to Prof. ZHANG Shuhuai for accepting them into his laboratory;
to Prof. OHKAWA Hiroshi, Assoc., for his insights regarding the presence of trichomes on the surfaces of tomato leaves;
and to Prof. Emeritus ARAKAWA Osamu from the Faculty of Agriculture and Life Science at Hirosaki University for his invaluable advice on experimental design.
Their contributions were critical to the success of this research.
We also thank Nigel Cattlin / Alamy Stock Photo for the image used in Figure \ref{fig2}.
Furthermore, we illustrated this manuscript via AI tools (OpenAI, DALL·E).
The authors reviewed and corrected the AI-generated images in Figure \ref{fig5} and Figure \ref{fig9} to ensure accuracy.
This research was conducted using resources of the United Graduate School of Agricultural Sciences at Iwate University and the Faculty of Agriculture and Life Science at Hirosaki University.
Part of this research has been accepted for publication in IEEE Access (DOI: 10.1109/ACCESS.2024.3500215).

\newpage

\printbibliography 

\end{document}